\documentclass{article}

\usepackage[T1]{fontenc}
\usepackage{textcomp}
\usepackage[final]{corl_2026}

\usepackage{amsmath,amssymb}
\usepackage{algorithm}
\usepackage{algorithmic}

\usepackage{graphicx}
\usepackage{float}
\usepackage{array}
\usepackage{wrapfig}
\usepackage{needspace}
\usepackage{subcaption}
\usepackage{stfloats}
\usepackage{booktabs}
\usepackage{makecell}
\usepackage{multirow}
\usepackage[table]{xcolor}
\usepackage{placeins}

\usepackage{makecell} 
\usepackage[misc]{ifsym}        
\usepackage{multirow}           
\usepackage{amsmath,amssymb}    

\usepackage{cleveref}
\usepackage{csquotes}
\usepackage{adjustbox}
\usepackage{url}
\usepackage{longtable}          
\usepackage{tabularx}           
\usepackage{hhline}

\usepackage{tikz}
\usetikzlibrary{shapes, arrows.meta, positioning}

\usepackage{hyperref}
\usepackage{nicematrix}
\usepackage{bm}
\usepackage{pifont}

\usepackage{minted}
\usepackage[skins, breakable]{tcolorbox}
\usepackage{listings}

\usepackage{wrapfig}

\newcommand{\eat}[1]{}                                 

\newcommand{\ourmethod}{{RoboGene}}
\newcommand{\methodname}{{RoboGene}}

\definecolor{bblue}{rgb}{0.85,0.92,1.0}
\newcommand{\bbluecell}[1]{\cellcolor{bblue}{#1}}

\title{
RoboGene: Diversity-Driven Agentic Task Generation Framework for VLA Pre-training}

\author{
{\bfseries
Yixue Zhang$^{1,3,*}$ \quad
Kun Wu$^{1,*}$ \quad
Zhi Gao$^{4}$ \quad
Zhen Zhao$^{1}$ \quad
Pei Ren$^{1}$
}\\
{\bfseries
Zhiyuan Xu$^{1}$ \quad
Fei Liao$^{1}$ \quad
Xinhua Wang$^{1}$ \quad
Shichao Fan$^{1}$ \quad
Di Wu$^{1,5}$
}\\
{\bfseries
Qiuxuan Feng$^{1,5}$ \quad
Meng Li$^{1}$ \quad
Zhengping Che$^{1,\dagger,\ddagger}$ \quad
Chang Liu$^{2,\ddagger}$ \quad
Jian Tang$^{1,\ddagger}$
}\\[0.5em]
\normalfont
$^{1}$Beijing Innovation Center of Humanoid Robotics,
Beijing, China\\
$^{2}$School of Advanced Manufacturing and Robotics,
Peking University, Beijing, China\\
$^{3}$School of Mechanics and Engineering Science,
Peking University, Beijing, China\\
$^{4}$Beijing Institute of Technology,
Beijing, China\\
$^{5}$State Key Laboratory of Multimedia Information Processing,\\
School of Computer Science, Peking University,
Beijing, China\\[0.5em]
\texttt{z.che@x-humanoid.com, changliucoe@pku.edu.cn,}\\
\texttt{jian.tang@x-humanoid.com}\\[0.3em]
$^{*}$Equal contribution.
\quad
$^{\dagger}$Project lead.
\quad
$^{\ddagger}$Corresponding authors.
}

\begin{document}
\raggedbottom
\maketitle

\begin{abstract}
The pursuit of general-purpose robotic manipulation is hindered by the scarcity of diverse, real-world interaction data.
Unlike the large-scale web vision-language data, robotic data collection is an active process incurring prohibitive physical costs.
As a result, the key challenge is not merely to collect more robot data, but to obtain training data that provides informative learning knowledge by broadening behavioral coverage and capturing meaningful variations in objects, scenes, and skills. However, \textbf{how to automatically curate task proposals that maximize the utility of collected robot data remains a critical yet under-explored challenge.}
To address this, we introduce \textbf{\methodname}, an agentic framework designed to automate the generation of diverse and informative manipulation tasks across heterogeneous robotic embodiments, spanning single-arm, dual-arm, and mobile robots. 
\methodname~integrates three core components: (1) diversity-driven sampling for broad task coverage, (2) agentic self-reflection for task-quality assessment, and (3) human-verified feedback memory for execution-grounded validation and continual refinement.
We conduct extensive quantitative analysis and large-scale real-world experiments, collecting datasets of 18k trajectories and introducing novel metrics to assess task quality in terms of feasibility and diversity. 
Results demonstrate that \methodname~significantly outperforms frontier foundation models (e.g., GPT-4o, Gemini 2.5 Pro). 
Furthermore, real-world experiments show that VLA models trained on data generated by \methodname~achieve higher success rates and stronger generalization, highlighting the importance of high-quality task generation.
Our project page and dataset are publicly available at
\href{https://robogene-boost-vla.github.io/}{robogene-boost-vla.github.io},
with dataset repositories hosted on
\href{https://huggingface.co/datasets/X-Humanoid/RoboGene}{Hugging Face}
and
\href{https://modelscope.cn/datasets/X-Humanoid/RoboGene_tasks}{ModelScope}.

\end{abstract}
\keywords{Robot Learning, Vision-Language-Action Models, Task Generation}
\section{Introduction}
\begin{figure}[!t]
    \centering
    \includegraphics[
        width=0.98\linewidth,
        trim=1mm 2mm 2mm 0mm,
        clip
    ]{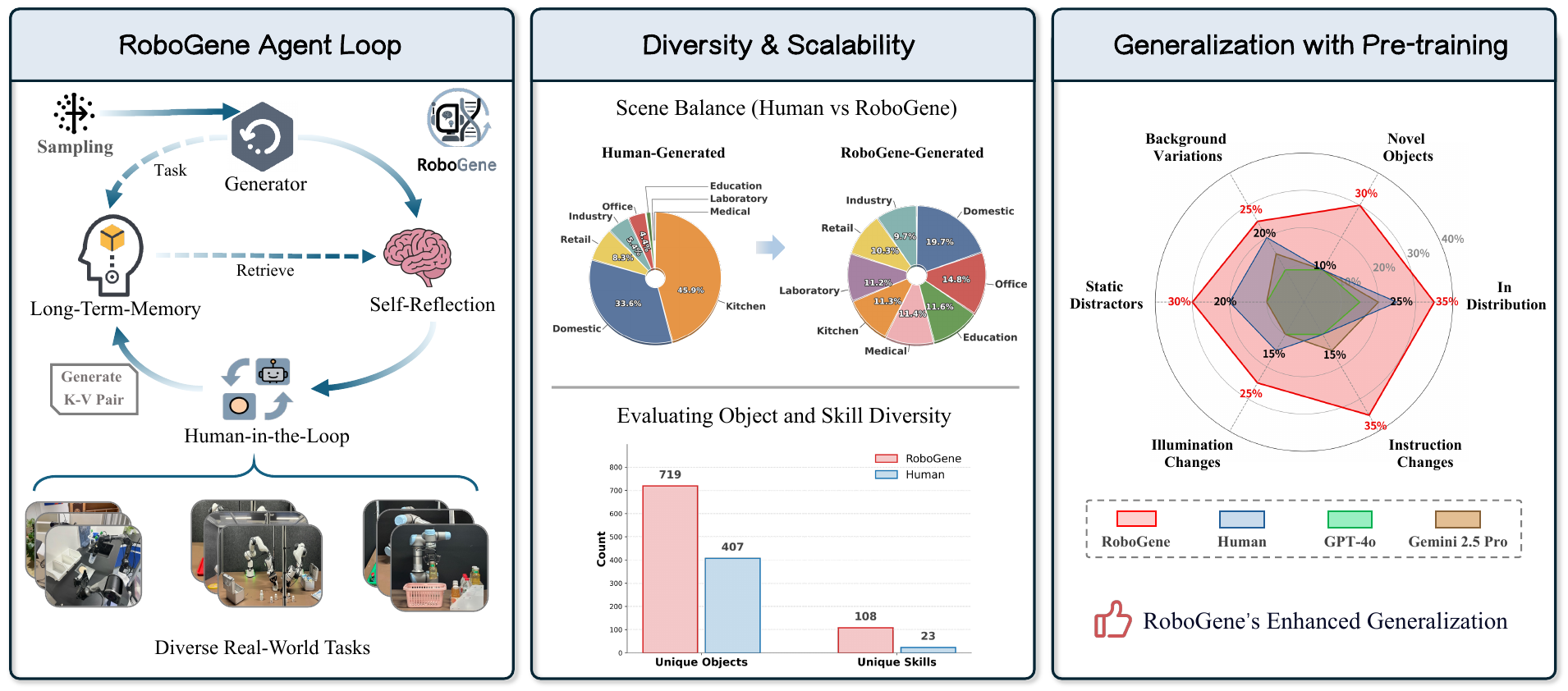}
    \vspace{-0.6em}
    \caption{RoboGene is an agentic framework designed to automate the generation of diverse, informative, and physically plausible manipulation tasks, thereby enhancing the foundational capability and generalizability of VLA model pre-training.}
    \label{fig:motivation}
    \vspace{-15pt}
\end{figure}


Foundation models trained on web-scale data have shown strong generalization across diverse tasks~\cite{pfeifer2004embodied,firoozi2025foundation}, motivating the community to build general-purpose embodied agents for manipulation in unstructured environments~\cite{zitkovich2023rt,kim2024openvla,liu2024rdt,cheang2025gr,lee2025molmoact,intelligence2025pi_05,bu2025univla}. This goal hinges on large datasets that equip models with broad physical knowledge and skills~\cite{o2024open,wu2024robomind,hou2025robomind}, yet robot data, unlike text or images, must be gathered through costly physical interaction. Recent efforts mainly scale hardware to raise data volume, while the question of which tasks to collect receives far less attention. The prevalent approach~\cite{o2024open,walke2023bridgedata,khazatsky2024droid} relies on human operators to specify tasks by hand, which is hard to scale and prone to cognitive bias. Designers favor simple primitives such as ``pick the apple'' over complex, multi-stage interactions, leaving the data severely long-tailed and limiting generalization to novel instructions and scenes. This raises a central question. \textit{How can we automate the design of high-quality and diverse collection tasks to improve pre-training and generalist capability?}

A natural remedy is to use Large Language Models (LLMs) for automated task generation~\cite{wang2024robogen,wang2023gensim,ha2023scaling,huagensim2,dai2025manitaskgen,gong2025anytask}, which improves scalability but raises new obstacles. First, without access to global dataset statistics, an LLM cannot reshape the skewed distribution toward a more balanced one, and therefore fails to correct the long-tail effect. Second, LLMs lack physical grounding and are prone to hallucination, often producing tasks that violate user constraints or reference non-existent objects. They also cannot judge whether a given robot embodiment is able to execute a generated instruction, so guaranteeing consistency and feasibility still demands heavy expert effort. Finally, the generation is typically open-loop and ignores real-world execution feedback, such as object substitution or motion failure, that could refine later generations.

In this paper, we present \textbf{\ourmethod}, an agentic framework for the automated, large-scale generation of diverse, physically grounded, and high-quality manipulation tasks. To counter the long-tail problem, we adopt a diversity-driven sampling scheme based on a Least Frequently Used (LFU) strategy that steers the agent toward under-explored parts of the task space and prioritizes rare objects and skills. To suppress hallucination, a self-reflection module routes each task proposal through three specialized evaluators that independently check constraint adherence, novelty, and embodied plausibility before human verification. A long-term memory module further consolidates Human-in-the-Loop (HITL) feedback, assimilating corrections from task edits and execution failures during real-world collection, thereby enabling subsequent task generation to produce more robot-executable tasks. Together, these components substantially reduce expert labor while keeping the resulting task distribution both executable and statistically balanced.

We validate \ourmethod~through quantitative analysis and large-scale real-world experiments. We design a suite of metrics that capture both per-task quality, such as physical feasibility, and global dataset properties, such as object and skill diversity. \ourmethod~clearly outperforms strong foundation models such as GPT-4o~\cite{hurst2024gpt} and Gemini 2.5 Pro~\cite{comanici2025gemini}, as well as manual design. For real-world validation, we collect over 18k trajectories spanning 1200 distinct tasks with 15 demonstrations each to benchmark VLA pre-training. A $\pi_{0}$~\cite{black2025pi0} policy pre-trained on \ourmethod-generated tasks generalizes better to unseen scenarios, including novel objects, background changes, static distractors, lighting shifts, and new instructions. These results confirm that the diversity induced by \ourmethod~is critical for pre-training generalist policies. Our main contributions are summarized as follows:

\begin{itemize}
    \item We identify the lack of diversity and physical grounding in task generation as a fundamental barrier to learning generalist policies.
    \item We propose \ourmethod, an agentic framework that automates task generation through diversity sampling, self-reflection, and memory consolidation.
    \item We provide extensive empirical evidence showing that our generated data facilitates robust policy learning.
    \item We open-source a high-quality, large-scale real-world manipulation dataset of RoboGene-generated tasks to support future research.
\end{itemize}

\section{Related Work}

\subsection{Generalist Vision-Language-Action Models}

Large-scale robotic datasets~\cite{o2024open,wu2024robomind,hou2025robomind,walke2023bridgedata,khazatsky2024droid,ebert2021bridge,bu2025agibot,wu2025robocoin} have made it possible to move beyond specialized manipulation policies~\cite{chi2023diffusion_policy,bharadhwaj2024roboagent,wu2024swbt,buamanee2024bi,wu2024discrete,fu2024mobile,ze2024DP3,xia2026robovaluerlreliablevalueestimation} toward generalist Vision-Language-Action (VLA) models~\cite{zitkovich2023rt,liu2024rdt,black2025pi0,wen2025tinyvla}. 
Recent work further improves VLA architectures with stronger open-world generalization, diffusion or autoregressive action modeling, multisensory grounding, spatial reasoning, and cross-embodiment representations~\cite{intelligence2025pi_05,wen2025diffusionvla,liu2025mla,qu2025spatialvla,zheng2025x,fan2025xr}. 
Complementary to model-side progress, data-centric studies show that generalization depends not only on the amount of robot data, but also on its coverage across environments, objects, and tasks~\cite{lin2025datascalinglaws,shi2026diversity}. 
In particular, increasing demonstrations for the same task can yield diminishing returns, whereas broader task and environment diversity can improve downstream adaptation. 
These findings shift the key bottleneck from simply collecting more trajectories to deciding which tasks should be collected under limited and expensive robot time. 
However, existing large-scale datasets are still largely shaped by human-specified tasks and therefore inherit long-tail biases toward common objects, scenes, and primitive skills. 
\ourmethod~addresses this data-utility problem by automatically generating diverse, physically grounded task specifications for VLA pre-training.

\subsection{Automated Task Generation}

Automated task generation has evolved from procedural task templates~\cite{shridhar2020alfred} to LLM-driven systems that generate task descriptions, simulation environments, and reward functions~\cite{wang2024robogen,wang2023gensim,ha2023scaling,huagensim2,chen2025rcaregen}. 
This direction has been extended to curriculum design~\cite{ryu2025curricullm,hu2025agentgen}, realistic task and scene synthesis~\cite{nasiriany2024robocasa,ko2025gaia,zook2025grs,ye2025video2policy}, and sim-to-real data generation~\cite{dai2025manitaskgen,gong2025anytask,jiang2025dexmimicgen}. 
Despite these advances, most existing methods treat task generation as a component of simulation or benchmark construction, rather than as a closed-loop data curation problem for real-world VLA pre-training. 
They usually do not explicitly balance the global distribution of generated tasks, and open-loop generation can still produce hallucinated assets, incompatible object-skill pairs, or instructions that exceed the capability of a given robot embodiment. 
In contrast, \ourmethod~combines diversity-driven sampling, self-reflective feasibility checking, and human-verified execution feedback through long-term memory. 
This enables generated tasks to be not only semantically varied, but also executable and useful as real-world data-collection targets for pre-training generalist policies.

\begin{wrapfigure}[4]{r}{0.42\textwidth}
    \vspace{-18pt}
    \centering
    \includegraphics[
        width=0.40\textwidth,
        trim=110pt 315pt 585pt 95pt,
        clip
    ]{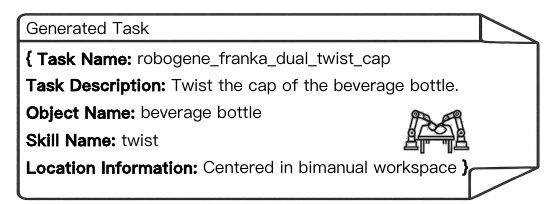}
    \vspace{-8pt}
    \caption{\small Main context of task example}
    \label{fig:task_example}
    \vspace{-11pt}
\end{wrapfigure}
\section{Methodology}
\label{sec:method}

\subsection{Overview}
\begin{figure*}[!t]
    \centering
    \includegraphics[width=0.95\linewidth]{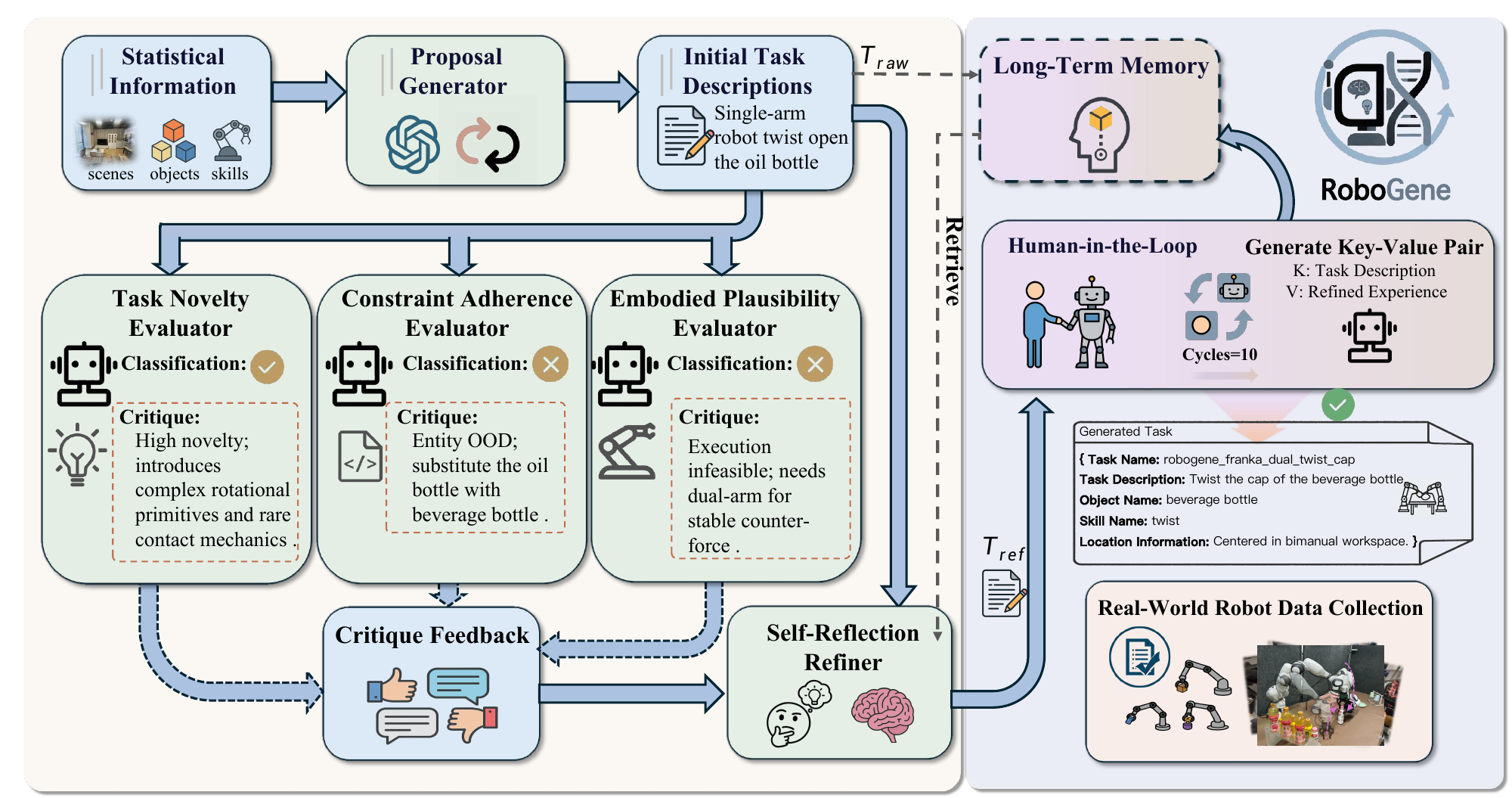}
    \caption{\textbf{Overview of RoboGene.} The framework comprises three key components: (1) a diversity-driven sampling mechanism based on a Least Frequently Used (LFU) strategy; (2) a self-reflection mechanism that enhances task generation quality; and (3) a long-term memory module that consolidates Human-in-the-Loop (HITL) feedback.}
    \label{fig:main_method}
    \vspace{-15pt}
\end{figure*}
We address the challenge of automating diverse task design for real-world robotic data collection. RoboGene operates on four predefined libraries: a robot-embodiment library ($\mathcal{R}$), a scenario list ($\mathcal{E}$), an object list ($\mathcal{O}$), and a manipulation-skill list ($\mathcal{S}$), which can be replaced or extended by users, whose example contents are provided in Appendix~\ref{sec:appendix_a_libraries}. 

Each generated task instance is defined as $\tau = (r_\tau,n_\tau, d_\tau, \ell_\tau, e_\tau, O_\tau, S_\tau, L_\tau, C_\tau)$, where $n_\tau$, $d_\tau$, and $\ell_\tau$ denote the task name, natural-language task description, and executable language instruction, respectively; $e_\tau \in \mathcal{E}$ denotes a predefined scenario category; $r_\tau \in \mathcal{R}$ denotes the target robot embodiment; $O_\tau \subseteq \mathcal{O}$ and $S_\tau \subseteq \mathcal{S}$ denote the selected objects and manipulation skills involved in the task; $L_\tau$ specifies the scene layout, including object placements and workspace assignments; and $C_\tau$ provides additional task context such as object poses, affordances. Our objective is to synthesize a set of physically valid and diverse task instances $\mathcal{T}=\{\tau_i\}_{i=1}^{N}$ to facilitate foundation-model pre-training. We present task instances generated by RoboGene in Appendix~\ref{app:task_instances}, and schematic illustration of a task example in Figure~\ref{fig:task_example}.

To achieve this, we introduce RoboGene, an agentic framework that synthesizes tasks through a closed-loop process involving diversity-driven sampling, self-reflective generation, and memory-augmented refinement integrating human feedback. As illustrated in Figure~\ref{fig:main_method}, the generation process $\Phi$ transforms user prompt inputs into concrete task specifications. 
The overall task generation can be formalized as:
\begin{equation}
    T = \Phi_{\text{refine}} \Big( \Phi_{\text{gen}} \big( \Phi_{\text{sample}}(\mathcal{E}, \mathcal{O}, \mathcal{S} \mid H), \mathcal{R} \big) \mid \mathcal{M} \Big).
\end{equation}
Here, \(\Phi_{\mathrm{sample}}\) selects a scenario and its associated candidate objects and skills from the predefined libraries, guided by historical usage statistics \(H\) to improve distributional coverage. The generator \(\Phi_{\mathrm{gen}}\) denotes a task-generation agent, which drafts an initial task proposal and improves it through evaluator feedback and self-reflection. Finally, the refiner \(\Phi_{\mathrm{refine}}\) improves the task using a long-term memory module \(\mathcal{M}\) that consolidates human feedback. We provide the detailed algorithmic workflow of RoboGene in Appendix~\ref{app:implementation}. \ourmethod~uses GPT-4o as the default backbone.

\subsection{Diversity-Driven Task Space Sampling}
\label{ssec:sampling}


To counter the long-tail task distribution and ensure balanced coverage, we adopt a Least Frequently Used (LFU) sampling strategy guided by historical statistics $\mathcal{H}$. We maintain a usage counter $u(\cdot)$  for every scenario, object, and skill, incrementing it whenever a valid task is generated, so that sampling progressively favors under-explored elements. Each generation cycle $t$ proceeds hierarchically: we first select the least-used scenario $e_t = \arg\min_{e \in \mathcal{E}} u(e)$, and then sample objects and skills conditioned on $e_t$. To avoid incoherent pairings such as folding a microwave in a factory, we retain only a relevant subset $\mathcal{O}_{e_t} \subset \mathcal{O}$ whose cosine similarity to the scenario description, measured by a text encoder, exceeds a threshold $\theta = 0.7$ calibrated on manually labeled scenario--object pairs. From this subset we draw a fixed-size candidate set $\mathcal{O}_t$ via LFU, and obtain the skill set $\mathcal{S}_t$ in the same manner. The resulting tuple $(e_t, \mathcal{O}_t, \mathcal{S}_t)$ is then inserted into the prompt as context, since providing a candidate set rather than a single forced choice allows the agent to preserve diversity.

\subsection{Self-Reflection and Task Generation }
\label{ssec:reflection_generation}

Upon determining the sampling constraints $(e_t, O_t, S_t)$, the framework enters a generator-evaluator-improvement loop to produce a high-quality task specification $T$. We present prompt templates for each node in Appendix~\ref{app:prompts}.

\textbf{Task Proposal Generation.}
The initial task $T_{\text{raw}}$ is synthesized by a Proposal Generator $G(\cdot)$, implemented via a Large Language Model (LLM). 
We construct a structured prompt injecting the sampled constraints, robot type $r$, and agent role definitions. 
For embodiments requiring spatial awareness, such as mobile manipulators, $G(\cdot)$ is instantiated as a Vision-Language Model (VLM). 
This allows the system to process scene images $\{I_1, \dots, I_k\}$ as supplementary input, grounding the generated task instructions in the physical layout to ensure spatial plausibility. 
The output $T_{\text{raw}}$ is formatted as a standardized JSON containing the robot type, scenario category, task description, object, and skill lists.

\paragraph{Multi-Faceted Self-Reflection.}
Because the raw proposal $T_{\text{raw}}$ may contain hallucinations or infeasible steps, we route it through a set of specialized LLM-based evaluators. Each returns a natural-language critique, and a refiner then merges these critiques into a revised task.

The Embodied Plausibility Evaluator $E_{\text{emb}}(\cdot)$ judges whether the task can be carried out given the robot embodiment $r$, checking the consistency among robot capabilities, object affordances, and action order. It flags embodiment-specific issues, such as a single arm attempting a task that needs a stabilizing hand, two arms taking incoherent roles, or a mobile robot acting before reaching the target. When a step is infeasible or ambiguous, it suggests a fix, for example clarifying arm roles, adding stabilization, or splitting a coupled action. We denote its critique by $f_{\text{emb}} = E_{\text{emb}}(T_{\text{raw}}, r)$.

The Novelty Evaluator $E_{\text{nov}}(\cdot)$ rates difficulty and novelty along four dimensions, namely object complexity (e.g., deformable objects or fluids), contact precision (e.g., in-hand manipulation or tool use), temporal and logical depth (e.g., long-horizon tasks with partial observability), and environment structure (e.g., dynamic targets or disturbances). Its feedback $f_{\text{nov}} = E_{\text{nov}}(T_{\text{raw}})$ ensures that the task is sufficiently novel.

The Constraint Adherence Evaluator $E_{\text{con}}(\cdot)$ checks that the task matches the assigned scene and uses only objects and skills from the sampled sets $\mathcal{O}_t$ and $\mathcal{S}_t$, without introducing hallucinated items. Its feedback $f_{\text{con}} = E_{\text{con}}(T_{\text{raw}}, \mathcal{O}_t, \mathcal{S}_t)$ suggests edits to the task context.
Finally, the Self-Reflection Refiner $E_{\text{ref}}(\cdot)$ merges the three critiques with the raw proposal to refine.
\begin{equation}
T_{\text{ref}} = E_{\text{ref}}\!\left(T_{\text{raw}},\, \{f_{\text{emb}},\, f_{\text{nov}},\, f_{\text{con}}\}\right).
\end{equation}

\subsection{Integrating Human Feedback via Memory}
\label{ssec:ltm}

To enable continuous system evolution and prevent the recurrence of errors, we integrate a Human-in-the-Loop (HITL) mechanism coupled with a semantic Long-Term Memory module $\mathcal{M}$.

\textbf{Human-in-the-Loop Feedback Consolidation.}
When tasks are dispatched for real-world execution, human operators provide feedback not just as binary labels, but as rich natural language explanations detailing infeasibility reasons (e.g., "writing number '2' fails on opposite-facing arms due to spatial mapping inversion"). 
We employ an LLM-based summarizer to periodically distill these specific feedback instances into generalized heuristic knowledge which is stored in $\mathcal{M}$ as key-value pairs $(K, V)$, where the key is the semantic embedding of the task context and the value is the actionable guideline.

\textbf{Memory-Augmented Refinement.}
During the refinement phase, the system retrieves relevant knowledge from $\mathcal{M}$ to augment the self-reflection critiques. 
Specifically, we compute the embedding of the current task proposal and retrieve the top-$k$ most similar heuristic guidelines via cosine similarity. 
These retrieved insights are injected into the context window of the refiner $E_{ref} (\cdot)$. 
This Retrieval-Augmented Generation (RAG) approach~\cite{lewis2020retrieval} allows the agent to dynamically access a growing knowledge base of physical constraints and manipulation strategies, significantly enhancing the quality of generated tasks over time.
We present a concrete case study of this memory consolidation and retrieval process in Appendix~\ref{app:memory}.

\section{Experiments}
\label{sec:experiments}

We evaluate \ourmethod~by addressing three pivotal research questions regarding the validity, diversity, and utility of the generated tasks. 
Specifically, we investigate: 
(1) whether individual tasks are linguistically comprehensible and physically executable; 
(2) if the collective dataset exhibits sufficient coverage across scenarios, objects, and skills; 
and (3) the effectiveness of the collected data for downstream policy pre-training. 
To this end, we benchmark \ourmethod~against human experts, the rule-based method, and state-of-the-art Large Foundation Models (LFMs), specifically GPT-4o and Gemini 2.5 Pro. 
For each method, we generated a suite of 900 tasks spanning single-arm, dual-arm, and mobile manipulation domains, comprising 300 tasks for each robot category.
We provide the implementation details for the baseline methods in Appendix~\ref{app:baselines}.



\subsection{Individual Task Evaluation}

\begin{table}[!t]
    \centering
    \caption{
    Individual-task evaluation results on the 900 tasks generated by each method.
    Task Clarity, Type Consistency, and Logical Validity are averaged over
    independent human and Claude Opus 5 evaluations, while Object Coverage,
    Skill Coverage, and Physical Feasibility are directly measured.
    Higher is better for all metrics.
    }
    \label{tab:single_task_results}

    \footnotesize
    \setlength{\tabcolsep}{6pt}
    \renewcommand{\arraystretch}{1.1}
    \setlength{\aboverulesep}{1pt}
    \setlength{\belowrulesep}{1pt}

    \begin{tabular}{lcccccc}
        \toprule
        \multirow{2}{*}{Method}
        & Task ($\uparrow$)
        & Type ($\uparrow$)
        & Logical ($\uparrow$)
        & Obj. ($\uparrow$)
        & Skill ($\uparrow$)
        & Phys. ($\uparrow$) \\
        & Clarity
        & Consist.
        & Validity
        & Cov.
        & Cov.
        & Feas. \\
        \midrule

        Rule-based
        & 0.5472
        & 0.3967
        & 0.4011
        & 0.2145
        & 0.4840
        & 0.4811 \\

        Human
        & 0.9595
        & 0.8345
        & 0.9561
        & 0.3580
        & 0.1769
        & 0.9478 \\

        GPT-4o
        & 0.7978
        & 0.8456
        & 0.7573
        & 0.3105
        & 0.2542
        & 0.7467 \\

        Gemini 2.5 Pro
        & 0.8889
        & 0.8761
        & 0.6695
        & 0.2172
        & 0.2458
        & 0.6790 \\

        \midrule

        \bbluecell{\textbf{RoboGene}}
        & \bbluecell{\textbf{0.9911}}
        & \bbluecell{\textbf{0.9845}}
        & \bbluecell{\textbf{0.9928}}
        & \bbluecell{\textbf{0.6323}}
        & \bbluecell{\textbf{0.9153}}
        & \bbluecell{\textbf{0.9899}} \\

        \bottomrule
    \end{tabular}
\end{table}

\begin{figure*}[t]
    \centering \includegraphics[width=\textwidth,trim=0 20 0 10,clip]{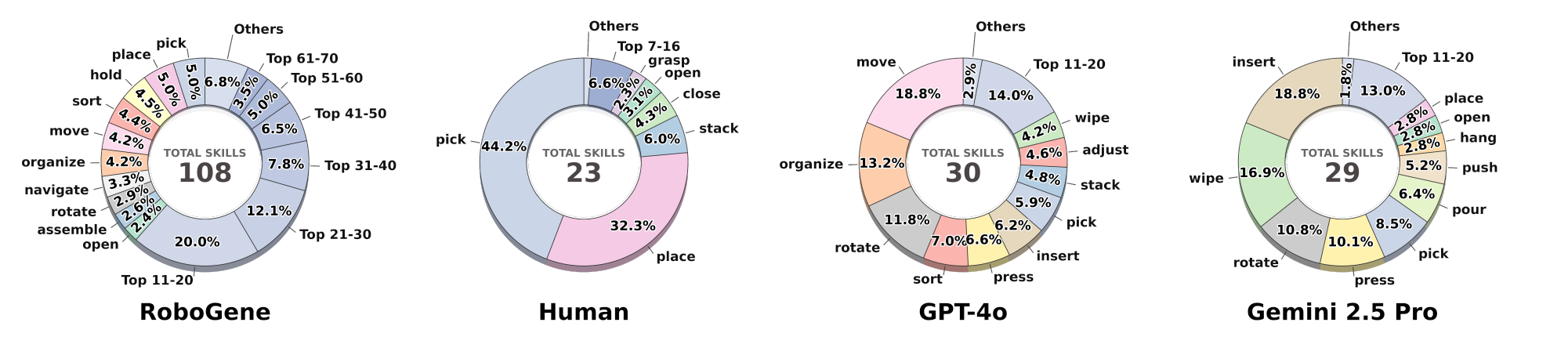}
    \caption{Skill diversity results given a total inventory of 118 skills. RoboGene significantly outperforms baselines in skill coverage and demonstrates a highly balanced distribution.}
    \label{fig:skill_diversity_results}
    \vspace{-15pt}
\end{figure*}

\textbf{Evaluation Metrics.} 
Quantifying the quality of open-ended robotic tasks remains an open challenge. 
We introduce six metrics to assess semantic and physical validity. 
\textit{Task Clarity} and \textit{Type Consistency} check for linguistic clarity and robot type alignment (e.g., dual-arm constraints). 
\textit{Logical Validity} ensures semantic plausibility, penalizing contextually inappropriate actions (e.g., industrial assembly in a kitchen). 
To measure grounding accuracy, we define \textit{Object Coverage} and \textit{Skill Coverage} as the ratio of generated entities present in the valid configuration sets $\mathcal{O}$ and $\mathcal{S}$, respectively. 
Finally, \textit{Physical Feasibility} reports the average success rate of tasks executed via human teleoperation (5 trials per task).
To reduce evaluator-specific bias, Task Clarity, Type Consistency,
and Logical Validity are independently assessed by human evaluators
and Claude Opus 5. We report the average in the main text and individual results in Appendix~\ref{app:metrics}.

\textbf{Evaluation Results.} 
Table~\ref{tab:single_task_results} demonstrates that \ourmethod~establishes a new state-of-the-art across all dimensions. 
While standard LFMs like GPT-4o and Gemini 2.5 Pro exhibit high task clarity, their coverage is fundamentally bottlenecked by severe hallucinations and limited long-tail diversity. Specifically, over 14.9\% of GPT-4o's and 52.0\% of Gemini's generated objects are physically absent from the valid library (see Appendix~\ref{app:metrics} for exact fraction details). 
In contrast, \ourmethod~effectively grounds all the generation within the available assets, achieving an object coverage of 0.6323 and a skill coverage of 0.9153. 
Furthermore, our self-reflection mechanism ensures physical grounding, outperforming the logical validity and physical feasibility of human-designed tasks, whereas rule-based baselines fail to produce coherent, physically executable behaviors.

\subsection{Dataset Diversity Analysis}
We further evaluate the distributional properties of the generated datasets, as diversity is critical for the generalization capability of VLA models. 
We analyze coverage across scenarios, objects, skills, and task semantic descriptions.
Additional implementation details for the ablation variants and
further experimental results are provided in
Appendices~\ref{app:ablation_details}
and~\ref{sec:add_exp}, respectively.

\begin{wrapfigure}[23]{r}{0.42\linewidth}
    \vspace{-0.8em}
    \centering

    \includegraphics[
        width=\linewidth,
        trim=5 5 5 5,
        clip
    ]{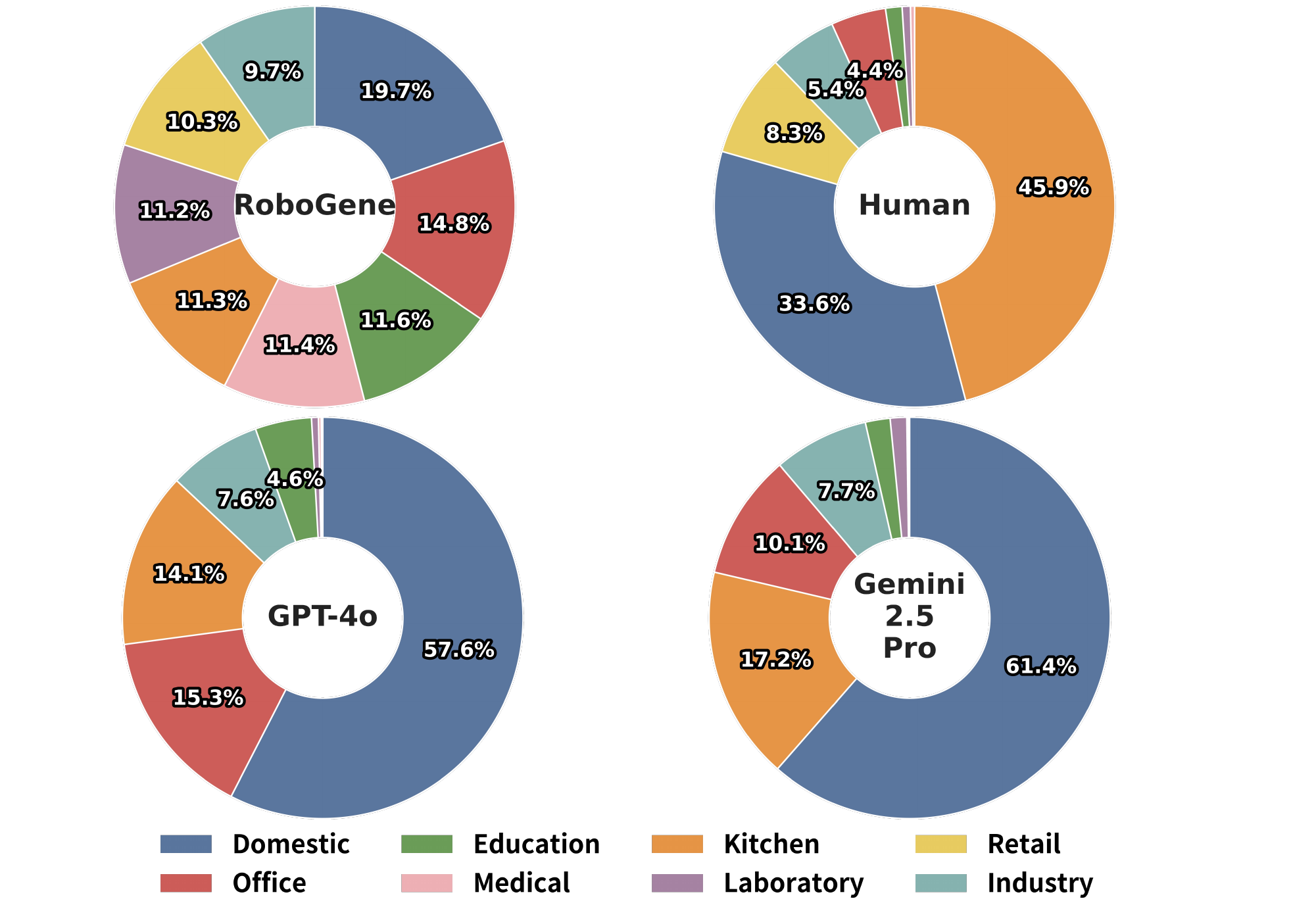}
    \vspace{-0.3em}

    {\footnotesize\textbf{(a)} Scenario diversity}
    
    \vspace{0.8em}

    \includegraphics[
        width=\linewidth,
        trim=5 5 5 5,
        clip
    ]{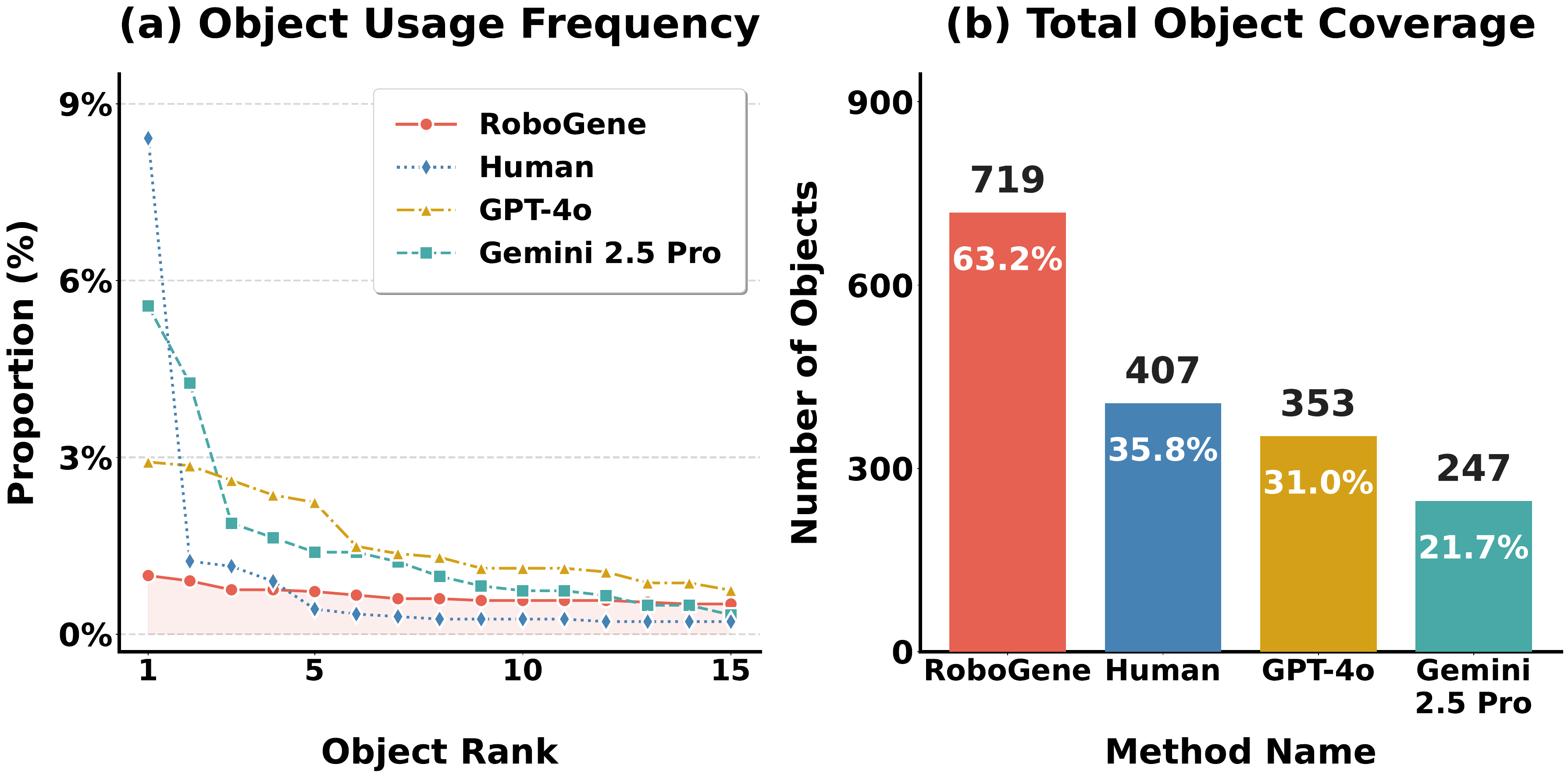}
    \vspace{-0.3em}

    {\footnotesize\textbf{(b)} Object diversity}

    \vspace{-0.5em}
    \caption{
    Diversity analysis of generated tasks.
    }
    \label{fig:diversity_analysis}
    \vspace{-1.0em}
\end{wrapfigure}
\textbf{Results on Skill Diversity.} 
We analyze the coverage of manipulation skills within the generated tasks in Figure~\ref{fig:skill_diversity_results}. 
\ourmethod~generates tasks encompassing 108 distinct skills, achieving a 91.5\% coverage of the skill space $\mathcal{S}$, substantially outperforming GPT-4o (25.4\%) and Gemini (24.6\%) baselines. 
The skill distribution in baseline methods is heavily skewed. 
For instance, simple primitives like pick and place dominate the human-designed dataset ($>$40\%). 
Conversely, \ourmethod~maintains a balanced distribution where top skills constitute less than 5\% each. 
This indicates our method successfully synthesizes complex, long-horizon behaviors (e.g., clean, arrange).

\textbf{Results on Scenario Diversity.} 
Figure~\ref{fig:diversity_analysis}(a) depicts the task distribution across eight semantic scenarios $\mathcal{E}$. 
LFMs exhibit a strong bias towards domestic settings, with GPT-4o concentrating nearly 90\% of tasks in home, kitchen, and office scenarios, likely reflecting biases in their pre-training corpora. 
\ourmethod~rectifies this imbalance, achieving a uniform distribution where no single category exceeds 20\%. 
By explicitly sampling under-represented environments (e.g., medical, industrial), our framework ensures the robotic agent encounters diverse physical constraints, from rigid industrial tools to delicate laboratory equipment.

\textbf{Results on Object Diversity.} 
Figure~\ref{fig:diversity_analysis}(b) illustrates the frequency distribution of objects in generated tasks. 
Human-curated datasets exhibit a sharp long-tail distribution, where a small set of dominant objects accounts for the majority of tasks, leaving the most objects under-represented. 
While direct usage of GPT-4o and Gemini 2.5 Pro partially mitigates this, they still suffer from ``generation saturation,'' plateauing at approximately 350 and 250 unique objects. 
\ourmethod~overcomes this bottleneck through a diversity-driven sampling strategy, covering 719 distinct objects, a $1.7\times$ expansion over the strongest baseline. 
The resulting distribution is flatter, ensuring that the VLA policy adapts to a broad spectrum of physical entities rather than overfitting to head classes.

\subsection{Real-World Manipulation Task Evaluation}

In this section, we evaluate the effectiveness of our pipeline through extensive real-world experiments. 
We first validate the feasibility of the generated tasks on physical robots. 
Subsequently, we assess the contribution of \ourmethod~to large-scale pre-training and its generalization capabilities.

\textbf{Validation of Generated Tasks.}
To verify the physical feasibility of our task generation pipeline, we conducted single-task evaluations on three distinct robots: a single-arm UR-5e, a dual-arm Franka, and an AgileX mobile robot. 
For each robot type, we generated three representative tasks using \ourmethod~and collected 250 trajectories per task. 
We train ACT~\cite{zhao2023act} for the stationary robots and $\pi_{0.5}$~\cite{intelligence2025pi_05} for the mobile robot and report the average success rates over 20 rollouts for each task. 
The system achieved robust performance, including a 90\% success rate on the SUR-SortButton task. 
The detailed experimental setups and quantitative success rates are deferred to Appendix~\ref{app:task_validation}.



\begin{figure}[t]
    \centering
    \includegraphics[width=0.8\linewidth]{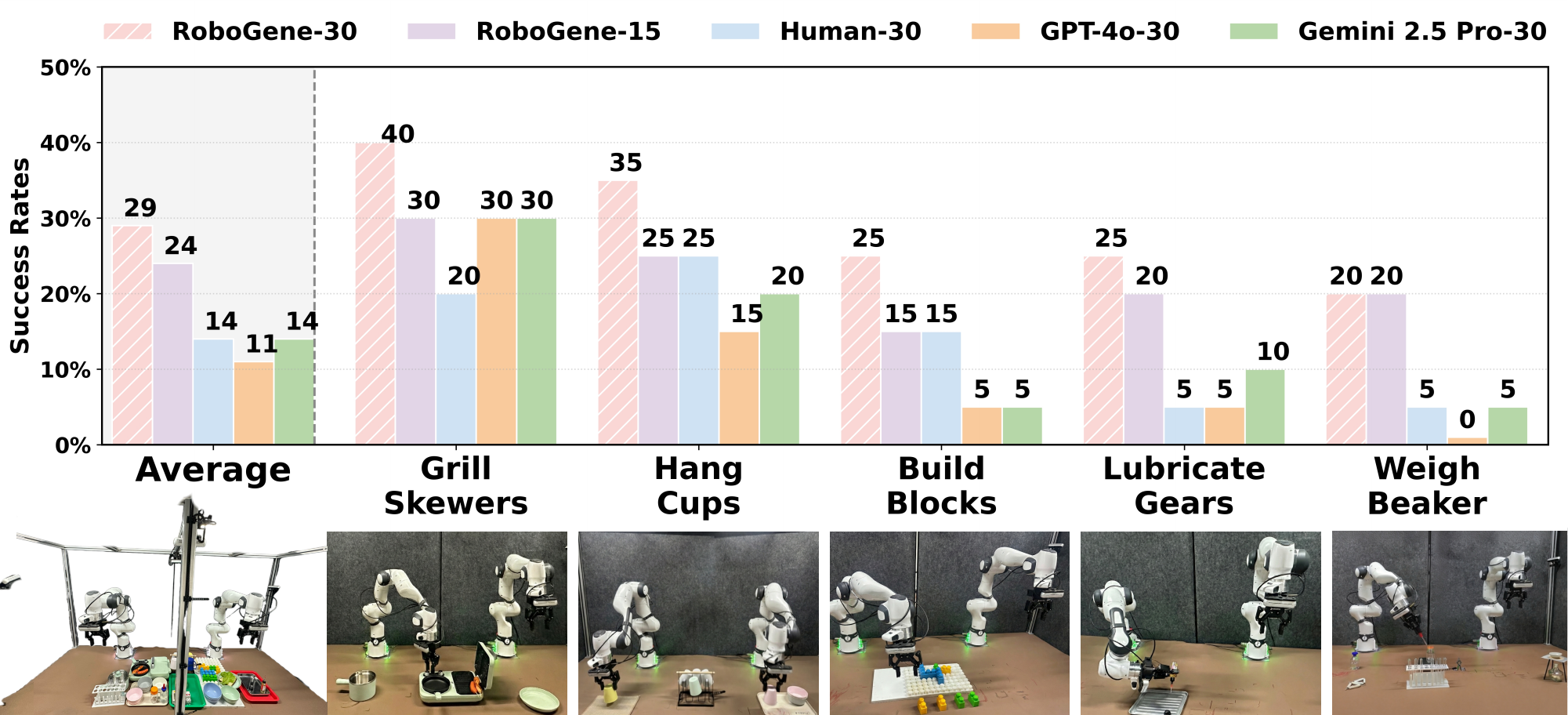}
    \caption{Success rates on 5 tasks after fine-tuning. The number appended to each method name indicates the number of fine-tuning epochs.}
    \label{fig:pretrain_multi-task}
\end{figure}

\textbf{Effectiveness of Pre-training Datasets}.
We constructed a dataset of 150 single-arm and 150 dual-arm tasks generated from each method to investigate the impact of large-scale pre-training on downstream performance. As shown in Figure~\ref{fig:pretrain_multi-task}, we pre-trained the $\pi_{0}$~\cite{black2025pi0} on these datasets and subsequently fine-tuned them on five unseen dual-arm tasks with 15 demonstrations per task, reporting the average success rate over 20 rollouts. 
Our results indicate a significant performance disparity between \ourmethod~and LLM-based baselines. 
Policies pre-trained on data designed by LFMs (GPT-4o, Gemini) struggled to adapt to new tasks, often yielding success rates below 20\% or failing completely. 
In contrast, 
the comparison between RoboGene-15 and 30-epoch baselines suggests that RoboGene pre-training enables faster downstream adaptation, reaching comparable or better performance with fewer fine-tuning epochs.

\subsection{Broader OOD Generalization}
\label{sec:broader_ood}
\begin{figure*}[t]
    \centering
    \includegraphics[width=\linewidth]{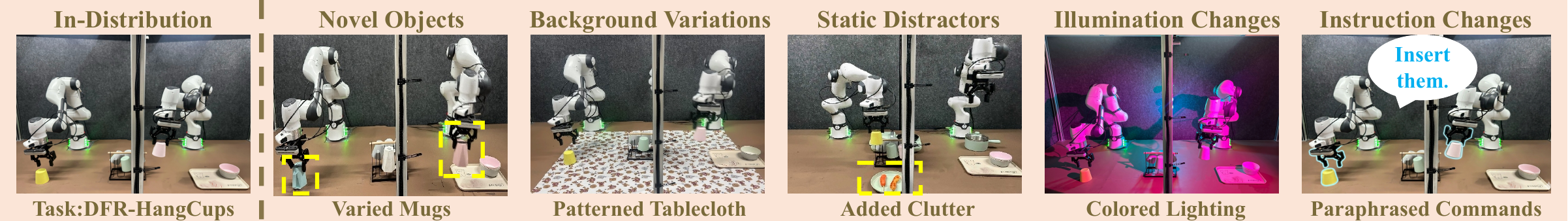}
    \caption{
Representative OOD shifts on the DFR-HangCups task, including
novel objects, background variations, static distractors,
illumination changes, and instruction changes.
The same five shift types are applied to all five downstream tasks
reported in Table~\ref{tab:broader_ood}.
}\label{fig:generalization_setting}
    \vspace{-15pt}
\end{figure*}
To evaluate the robustness of the learned policies, we conduct a broader OOD study on
five held-out dual-arm tasks.
For this study, each pre-trained policy is fine-tuned with
50 demonstrations per task and evaluated under five OOD shifts:
novel objects, background variations, static distractors,
illumination changes, and instruction changes.
Figure~\ref{fig:generalization_setting} illustrates the five OOD
shift types using DFR-HangCups as a representative example.
\begin{table}[t]
    \centering
    \caption{
    OOD generalization success rates (SR, \%) on five held-out
    dual-arm tasks. Each task aggregates 300 real-world trials
    across five OOD shifts.
    }
    \label{tab:broader_ood}

    \footnotesize
    \setlength{\tabcolsep}{5pt}
    \renewcommand{\arraystretch}{1.05}

    \begin{tabular}{@{}lcccccc@{}}
        \toprule
        Method
        & Grill
        & Hang
        & Build
        & Lubr.
        & Weigh
        & Avg. \\
        \midrule
        Human-designed
        & 42.0 & 15.0 & 2.0 & 24.0 & 12.0 & 19.0 \\
        GPT-4o
        & 30.0 & 8.0 & 6.0 & 26.0 & 9.0 & 15.8 \\
        Gemini 2.5 Pro
        & 46.0 & 12.0 & 7.0 & 17.0 & 7.0 & 17.8 \\
        \midrule
        \bbluecell{\textbf{RoboGene}}
        & \bbluecell{\textbf{67.0}}
        & \bbluecell{\textbf{40.0}}
        & \bbluecell{\textbf{25.0}}
        & \bbluecell{\textbf{48.0}}
        & \bbluecell{\textbf{21.0}}
        & \bbluecell{\textbf{40.2}} \\
        \bottomrule
    \end{tabular}
\end{table}

For each task and OOD shift, we conduct three runs of 20 real-world
trials, resulting in 300 evaluation trials per task.
As shown in Table~\ref{tab:broader_ood}, RoboGene achieves an
average success rate of 40.2\%, substantially outperforming
Human-designed tasks (19.0\%), GPT-4o (15.8\%), and
Gemini 2.5 Pro (17.8\%).
The improvement is consistent across all five tasks, indicating
that diversity-driven pre-training improves robustness beyond the
specific task and environment used for downstream adaptation.
Complete 95\% Wilson confidence intervals are provided in
Appendix~\ref{app:ood_ci}.











\subsection{Component Ablation}
\label{sec:component_ablation}

To isolate the contribution of each component, we conduct
remove-one-component ablations and include a balanced random
sampling baseline with constrained LLM generation.
For the first three metrics, we report the average of independent
human and Claude Opus 5 evaluations; the remaining metrics are
directly measured.
Rule-based (main-text Table~1) is library-only
\textbf{balanced-sampling baseline} without LLM. \textit{BRS+LLM} applies
constrained LLM generation to balanced-random library samples, without
reflection or memory. \textit{w/o Self-Reflection} retains LFU and memory but
removes iterative critique/refinement, serving as 
\textbf{single-pass constrained-prompting baseline}. LFU improves
coverage, Self-Reflection task correctness, and Memory physical feasibility. Additional analyses of human effort and memory accumulation
are provided in Appendix~\ref{sec:human_effort}.

\begin{table}[t]
    \centering
    \caption{
    Component ablation of RoboGene.
    Higher is better for all metrics.
    }
    \label{tab:component_ablation}

    \footnotesize
    \setlength{\tabcolsep}{4pt}
    \renewcommand{\arraystretch}{1.05}
    \setlength{\aboverulesep}{1.2pt}
    \setlength{\belowrulesep}{1.2pt}

    \begin{tabular}{@{}lcccccc@{}}
        \toprule
        \multirow{2}{*}{Method}
        & Task
        & Type
        & Logical
        & Obj.
        & Skill
        & Phys. \\
        & Clarity
        & Consist.
        & Validity
        & Cov.
        & Cov.
        & Feas. \\
        \midrule

        w/o LFU
        & 0.9884
        & 0.9784
        & 0.9900
        & 0.4494
        & 0.5000
        & 0.9867 \\

        w/o Self-Refl.
        & 0.8295
        & 0.8695
        & 0.7689
        & 0.5699
        & 0.7627
        & 0.9722 \\

        w/o Memory
        & 0.9906
        & 0.9811
        & 0.9650
        & 0.6025
        & 0.8136
        & 0.8500 \\

        BRS+LLM
        & 0.5778
        & 0.4556
        & 0.4778
        & 0.4667
        & 0.5300
        & 0.5133 \\

        \midrule

        \bbluecell{\textbf{RoboGene}}
        & \bbluecell{\textbf{0.9911}}
        & \bbluecell{\textbf{0.9845}}
        & \bbluecell{\textbf{0.9928}}
        & \bbluecell{\textbf{0.6323}}
        & \bbluecell{\textbf{0.9153}}
        & \bbluecell{\textbf{0.9899}} \\

        \bottomrule
    \end{tabular}
\end{table}
\section{Conclusion and Limitations}
\label{sec:conclusion}

We introduce \ourmethod, an agentic framework that automates the generation of diverse and physically grounded manipulation tasks, addressing the data diversity bottleneck in robotic learning. By pairing a diversity-driven LFU sampling strategy with a multi-faceted self-reflection mechanism, \ourmethod~ mitigates the hallucinations of large foundation models. Experiments show that it yields substantially higher object and skill coverage than state-of-the-art baselines, and trains VLA policies with stronger robustness to unseen objects, backgrounds, lighting, and instruction shifts. We open-source our resources to support future research.

\textbf{Limitations.}
\ourmethod~ relies on predefined scenario, object, and skill libraries, which enable controllable generation and systematic diversity sampling but also bound the task space to their coverage. Tasks involving unseen objects or skills thus require manual library expansion. Future work will explore automatic expansion from robot logs, scene observations, and web-scale knowledge, enlarging \ourmethod's task space while preserving physical grounding and controllable diversity.

\clearpage

\acknowledgments{

This work was supported by the Beijing Nova Program under Grant No.~20240484498, the National Natural Science Foundation of China under
Grant No.~92582205 and the Tianjin Municipal Science and
Technology Bureau under Grant No.~25YFXTHZ00340.

The large-scale real-world data collection and evaluation conducted
in this work relied on extensive collaboration across task setup,
robot teleoperation, data collection, quality assurance, and
experimental evaluation.
We sincerely thank Wenke Xia, Jianwei Guo, Shuguang Qiao, Jianwei Sun, Shiwei Jiao, Qichun Liu, Panpan Zhang, Tongze Liu, Qingyang Gao, Haonan Liu, Panpan Chen,  
and Tengfei Li for their valuable contributions to this project. Their collective efforts were essential to the completion of this work.

The completion of this work also relied on substantial infrastructure
support across the entire research pipeline, including robotic
hardware, system integration, data collection, quality assurance,
and experimental testing. We gratefully acknowledge the Beijing
Innovation Center of Humanoid Robotics for providing the robotic
platforms, experimental infrastructure, and operational support
essential to this work.

}
\bibliography{reference}

\clearpage
\appendix

\section{Implementation Details of RoboGene}
\label{app:implementation}
\paragraph{Framework Overview.}
RoboGene operates as a closed-loop agentic framework designed to autonomously generate high-quality, physically grounded, and diverse robotic manipulation tasks within real-world environments. 
The framework is founded upon three synergistic components devised to address the inherent challenges of automated data generation. 
First, we employ a diversity-driven sampling mechanism based on a Least Frequently Used (LFU) strategy to mitigate the long-tail distribution problem often observed in robotic datasets. 
This mechanism actively steers the generative agent toward under-explored regions of the task space, prioritizing interactions with rarely utilized objects and skills. 
Second, to improve the quality of the tasks, we incorporate a self-reflection mechanism. 
This module utilizes three specialized evaluators built on foundation models to independently assess constraint adherence, novelty, and Embodied Plausibility.
These evaluators rigorously scrutinize generated task proposals to provide actionable feedback. 
Third, the framework features a long-term memory module that consolidates Human-in-the-Loop (HITL) feedback. 
By assimilating corrections derived from task modifications and execution failures encountered during real-world collection, the system progressively refines its understanding of physical constraints and continuously enhances generation quality over time.
\paragraph{Task Space Definition.}
Formally, the generation process requires the specification of a robot embodiment, denoted as $r \in \mathcal{R}$, which is selected from three primary categories: single-arm robots, dual-arm robots, and mobile dual-arm robots. 
To define the semantic boundaries of the generation space, the system accepts a global scenario set $\mathcal{E}$, an object library $\mathcal{O}$, and a skill library $\mathcal{S}$. 
These definitions can be provided by the user as structured text files to align with specific hardware availability or data collection needs. 
\paragraph{Scenario and Library Configuration.}
In our specific implementation, we define $|\mathcal{E}| = 8$ distinct scenario categories, encompassing Domestic, Office, Education, Laboratory, Kitchen, Industry, Retail, and Medical environments. The scenario category \(e\) serves as a high-level physical context for task generation rather than a learned decision boundary. For single-arm and dual-arm manipulation, it corresponds to the tabletop workspace setup in which the objects are arranged and manipulated. For mobile manipulation, it corresponds to the physical environment in which the robot navigates and performs the task, such as a kitchen, office, warehouse, or medical room. These scenario categories are manually specified as part of the task-generation taxonomy and are used to constrain object and skill selection toward semantically coherent and physically executable tasks.
The object library $\mathcal{O}$ comprises 1,137 distinct physical entities, while the skill library $\mathcal{S}$ contains 118 manipulation primitives. 
This task space allows for high flexibility; users can customize the libraries based on their specific experimental setup or leverage Large Foundation Models (LFMs) to autonomously expand the semantic space. 
Additionally, RoboGene provides a remark interface for natural language constraints, allowing users to input specific requirements such as ``generate long-horizon tasks composed of 3-4 sub-goals'' or ``prioritize tasks requiring dexterous manipulation,'' which function as soft prompts during the generation phase.
\paragraph{Algorithmic Workflow.}
\begin{algorithm}[t]
\caption{RoboGene Iterative Task Generation Process}
\label{alg:robogene_flow}
\begin{algorithmic}[1]
\REQUIRE Robot type $r$, scenario set $\mathcal{E}$, object library $\mathcal{O}$, skill library $\mathcal{S}$, long-term memory $\mathcal{M}$
\STATE Initialize usage counters $u(e), u(o), u(s) \leftarrow 0$ for all $e \in \mathcal{E}$, $o \in \mathcal{O}$, $s \in \mathcal{S}$
\STATE Initialize history statistics $H \leftarrow \{u(e), u(o), u(s)\}$

\WHILE{generation loop is active}
    \STATE \textit{\textbf{\# Diversity-Driven Sampling}}
    \STATE Select target scenario $e_t \leftarrow \arg\min_{e \in \mathcal{E}} u(e)$ to prioritize under-explored environments
    \STATE Filter context-relevant subsets $\mathcal{O}_{e_t} \subset \mathcal{O}$ and $\mathcal{S}_{e_t} \subset \mathcal{S}$ based on semantic similarity to $e_t$
    \STATE Sample candidate sets $O_t \subset \mathcal{O}_{e_t}$ and $S_t \subset \mathcal{S}_{e_t}$ using the LFU strategy based on $H$
    
    \STATE \textit{\textbf{\# Task Proposal Generation}}
    \STATE Generate initial task proposal $T_{\mathrm{raw}} \leftarrow \Phi_{\mathrm{gen}}(r, e_t, O_t, S_t)$
    
    \STATE \textit{\textbf{\# Self-Reflection Evaluation}}
    \STATE $f_{\mathrm{nov}} \leftarrow E_{\mathrm{nov}}(T_{\mathrm{raw}})$ \COMMENT{Evaluate novelty and complexity}
    \STATE $f_{\mathrm{con}} \leftarrow E_{\mathrm{con}}(T_{\mathrm{raw}})$ \COMMENT{Verify adherence to sampled constraints}
    \STATE $f_{\mathrm{emb}} \leftarrow E_{\mathrm{emb}}(T_{\mathrm{raw}}, r, e_t, O_t, S_t)$ \COMMENT{Assess embodied task plausibility}
    
    \STATE \textit{\textbf{\# Memory-Augmented Refinement}}
    \STATE Retrieve relevant heuristic guidelines $v_{\mathrm{ret}} \leftarrow \operatorname{Retrieve}(\mathcal{M}, T_{\mathrm{raw}})$
    \STATE Refine task proposal $T_{\mathrm{ref}} \leftarrow \Phi_{\mathrm{refine}}(T_{\mathrm{raw}}, \{f_{\mathrm{nov}}, f_{\mathrm{con}}, f_{\mathrm{emb}}\}, v_{\mathrm{ret}})$
    
    \STATE \textit{\textbf{\# Human-in-the-Loop Verification}}
    \STATE Obtain verification feedback $b_t$ for $T_{\mathrm{ref}}$
    \IF{$b_t$ verifies $T_{\mathrm{ref}}$}
         \STATE Add $T_{\mathrm{ref}}$ to the generated task dataset
         \STATE Update usage counters and history statistics $H$
    \ELSE
         \STATE $v_{\mathrm{fail}} \leftarrow \Phi_{\mathrm{mem}}^{\text{GPT-4o}}(T_{\mathrm{ref}}, b_t)$ \COMMENT{Summarize reusable failure heuristic}
         \STATE $\mathcal{M} \leftarrow \mathcal{M} \cup \{(T_{\mathrm{ref}}, v_{\mathrm{fail}})\}$
    \ENDIF
\ENDWHILE
\end{algorithmic}
\end{algorithm}
The core execution logic of RoboGene is formalized in Algorithm~\ref{alg:robogene_flow}. 
The process is structured as an iterative loop that transitions through sampling, generation, reflection, and refinement phases to produce valid task instances. 
We initialize the system by setting usage counters $u(\cdot)$ and history statistics $H$ to zero. 
In the \textit{Diversity-Driven Sampling} phase, the system identifies the target scenario $e_t$ with the minimum usage count to prioritize under-represented environments. 
It then filters context-relevant objects $\mathcal{O}_{e_t}$ and skills $\mathcal{S}_{e_t}$ based on semantic similarity, subsequently sampling candidate sets $O_t$ and $S_t$ using the LFU strategy. 

Following sampling, the generator $\Phi_{gen}$ synthesizes an initial task proposal $T_{raw}$ based on the robot type $r$ and sampled constraints. 
This proposal immediately enters the \textit{Self-Reflection Evaluation} phase, where it is scrutinized by three evaluators to produce specific critiques: $f_{nov}$ for novelty, $f_{con}$ for constraint adherence, and $f_{emb}$ for Embodied Plausibility. 
In the \textit{Memory-Augmented Refinement} phase, the system retrieves relevant heuristic guidelines $v$ from the long-term memory $\mathcal{M}$. 
A refiner module $\Phi_{\mathrm{refine}}$ then synthesizes a finalized task
$T_{\mathrm{ref}}$ by integrating the original proposal,
the evaluator critiques, and the retrieved memory.
If $T_{\mathrm{ref}}$ passes human verification,
it is added to the generated task dataset and the history $\mathcal{H}$ is updated.
Otherwise, the corresponding execution feedback is summarized and consolidated
into the memory $\mathcal{M}$ to prevent similar failures in subsequent generations.
The resulting tasks are structured in a standardized JSON format, facilitating direct translation into execution code for both simulation and real-world deployment.
\paragraph{Data Usage Across Experiments.}

To avoid ambiguity among the different task sets and trajectory collections used in our experiments, we provide a complete accounting of the generated task specifications, pre-training data, downstream fine-tuning data, and real-world validation data in Table~\ref{tab:data_accounting}. 
\begin{table*}[t]
    \centering
    \caption{
        Dataset accounting and experimental splits.
        ``Task specifications'' denote generated task descriptions
        before trajectory collection.
        ``Trajectories'' denote real-world demonstrations or policy
        rollouts used for training or evaluation.
    }
    \label{tab:data_accounting}

    \footnotesize
    \setlength{\tabcolsep}{4pt}
    \renewcommand{\arraystretch}{1.05}
    \setlength{\aboverulesep}{1.2pt}
    \setlength{\belowrulesep}{1.2pt}

    \begin{tabularx}{0.98\textwidth}{@{}
        >{\raggedright\arraybackslash}X
        >{\centering\arraybackslash}X
        >{\centering\arraybackslash}X
        >{\centering\arraybackslash}p{0.105\textwidth}
        >{\centering\arraybackslash}X
        >{\centering\arraybackslash}X
    @{}}
        \toprule

        Purpose
        & Scope
        & Methods
        & Tasks
        & Data per Task
        & Total Data \\

        \midrule

        Task-quality and diversity evaluation
        & Single-arm, dual-arm, mobile
        & Rule-based, Human, GPT-4o, Gemini, RoboGene
        & 900 / method
        & Task specs only
        & 4,500 task specs \\

        \addlinespace[0.6em]

        VLA pre-training
        & Single-arm, dual-arm
        & Human, GPT-4o, Gemini, RoboGene
        & 300 / method
        & 15 demos
        & 18,000 trajectories \\

        \addlinespace[0.6em]

        Single-task real-world validation
        & UR-5e, dual-arm Franka, AgileX
        & RoboGene
        & 3 tasks / robot type
        & 250 demos
        & 2,250 trajectories \\

        \addlinespace[0.6em]

        Downstream fine-tuning
        & Five unseen dual-arm tasks
        & Pre-trained policy variants
        & 5 tasks
        & 15 demos / task
        & 75 demos / variant \\

        \addlinespace[0.6em]

        Generalization evaluation
        & Five held-out dual-arm tasks under five OOD shifts
        & Pre-trained policy variants
        & 5 tasks $\times$ 5 shifts
        & 50 fine-tuning demos/task;\newline
          3 runs $\times$ 20 trials/shift
        & 1,500 evaluation trials / method \\

        \bottomrule
    \end{tabularx}
\end{table*}
\paragraph{Task selection for VLA pre-training.}
For each generation method, we first generate 900 task specifications, balanced across three robot embodiments: 300 single-arm tasks, 300 dual-arm tasks, and 300 mobile-manipulation tasks. The task-quality and diversity evaluation is conducted on all 900 task specifications for each method. For the VLA pre-training comparison, we focus on tabletop single-arm and dual-arm manipulation, consistent with our downstream fine-tuning and generalization evaluations. Therefore, mobile-manipulation tasks are included in the task-generation and feasibility analysis, but are not used for the $\pi_0$ pre-training comparison.

From the single-arm and dual-arm task pools, we construct a balanced pre-training subset using stratified random sampling, selecting 150 single-arm tasks and 150 dual-arm tasks for each method. Task specifications that cannot be instantiated in our physical workspace due to unavailable objects, unsupported skills, or embodiment-inconsistent requirements are retained for the task-generation evaluation but excluded from trajectory collection. After applying the same feasibility-filtering protocol to all methods, we sample the pre-training subset from the remaining executable single-arm and dual-arm task pools. Importantly, all downstream fine-tuning and generalization tasks are held out from the pre-training task pools of all four methods, ensuring that no downstream evaluation task appears in any pre-training dataset. Each selected task is collected with 15 teleoperated demonstrations, yielding $4 \times 300 \times 15 = 18{,}000$ trajectories in total.

\section{Predefined Libraries of Scenarios, Objects, Skills and Robot Embodiment}
\label{sec:appendix_a_libraries}

As mentioned in the main text, RoboGene operates on four predefined libraries: a robot-embodiment
library $\mathcal{R}$, a scenario list $\mathcal{E}$, an object library
$\mathcal{O}$, and a manipulation-skill library $\mathcal{S}$, which can be replaced or extended by users. In our current implementation, these libraries encompass a total of 8 scenarios, 118 manipulation skills, and 1,137 distinct physical objects. Below, we provide a representative subset of these libraries to illustrate the diversity of our assets.

\subsection{Scenario List ($\mathcal{E}$)}
The scenario list contains 8 fundamental environments designed to cover a broad spectrum of real-world robotic applications:
\begin{itemize}
    \item Domestic, Education, Industry, Kitchen, Laboratory, Medical, Office, Retail.
\end{itemize}

\subsection{Manipulation-Skill List ($\mathcal{S}$)}
The full library contains 118 atomic and compositional manipulation skills. The following 10 representative skills demonstrate the varied kinematic and physical requirements handled by the generation engine:
\begin{itemize}
    \item \texttt{close}, \texttt{open}, \texttt{wipe}, \texttt{sort}, \texttt{stack}, \texttt{insert}, \texttt{adjust}, \texttt{assemble}, \texttt{fold}, \texttt{scoop}.
\end{itemize}
\subsection{Object List ($\mathcal{O}$)}
The global object list contains a massive inventory of 1,137 everyday and specialized items. To demonstrate the semantic and physical diversity of our assets, we list 4 representative objects for each of the 8 predefined scenarios:

\begin{itemize}
    \item \textbf{Domestic:} White clothes hanger, Slippers, White trash can, Green broom and dustpan.
    \item \textbf{Education:} Whiteboard marker, Whiteboard eraser, English letter building blocks, Early education toy blocks.
    \item \textbf{Industry:} Claw hammer, Phillips screwdriver, Emergency stop button switch, Industrial pure iron Q235.
    \item \textbf{Kitchen:} Frying pan, White cutting board, Grey IKEA bowl, Multifunctional cooking machine.
    \item \textbf{Laboratory:} Measuring cup, Large transparent funnel, White electronic scale, Transparent pump bottle.
    \item \textbf{Medical:} First-aid kit, Medical bandage, Adhesive bandage, Medical alcohol.
    \item \textbf{Office:} Black pen holder, Black stapler, A4 clipboard folder, Three-tier file storage rack.
    \item \textbf{Retail:} Shopping basket, QR code scanner, Canned Coca-Cola, Lay's potato chips.
\end{itemize}

\subsection{Robot Embodiment List ($\mathcal{R}$)}
The robot embodiment list contains three major categories used in our task-generation framework: single-arm manipulation, dual-arm manipulation, and mobile manipulation. In our implementation, these categories are instantiated with representative real-world robot platforms:
\begin{itemize}
    \item Single-arm manipulators: Universal Robots UR5e.
    \item Dual-arm manipulators: a dual-arm setup with two Franka Research 3 robotic arms.
    \item Mobile dual-arm manipulators: AgileX Cobot Magic V2.0 and ARX LIFT.
\end{itemize}

\section{Detailed Case Study of Memory-Augmented Refinement}\label{app:memory}

To provide a transparent overview of the RoboGene evolutionary loop, this section presents a structured case study tracing a complete failure-to-success lifecycle. The pipeline demonstrates how colloquial operational data-collection logs are converted into structural heuristic constraints within the long-term memory $\mathcal{M}$, and subsequently retrieved to rectify physical hallucinations.

\begin{center}
    \small\bfseries RoboGene Evolutionary Data Pipeline Workflow
\end{center}

\noindent\textbf{Phase 1: Initial Open-Loop Generation ($T_{raw}$)} \\
The Proposal Generator synthesizes a raw task targeting a stationary bimanual arm, blindly requesting fine-grained trajectory tracing without environmental mapping context.

\begin{tcolorbox}[title=Proposal Generation Output ($T_{raw}$), colback=blue!5, colframe=blue!40, fonttitle=\bfseries, breakable, before skip=4pt, after skip=4pt]
\begin{lstlisting}[
    basicstyle=\ttfamily\scriptsize, 
    breaklines=true,                 
    columns=fullflexible,            
    breakatwhitespace=false,         
    keepspaces=true                  
]
{
  "task_name": "robogene_franka_write_digit",
  "task_description": "The robot uses a marker to write the number '2' precisely on the tabletop whiteboard.",
  "language_instruction": "Pick up the marker and precisely write the number '2' on the whiteboard.",
  "objects": ["whiteboard marker", "tabletop whiteboard"],
  "skills": ["write"],
  "scene_layout": {
    "whiteboard marker": "Right Arm Workspace: Upper Right",
    "tabletop whiteboard": "Center of the table"},
  "task_context": "The whiteboard is fixed flat on the table facing the opposite operator's station."
}
\end{lstlisting}
\end{tcolorbox}

\begin{center} \Large $\Downarrow$ \end{center}

\noindent\textbf{Phase 2: Human Execution Evaluation \& LLM Distillation ($b_t$)} \\
The task is dispatched to the physical deployment pipeline. The teleoperation operator attempts execution, identifies a severe perspective conflict, and enters colloquial log notes. The LLM-based summarizer immediately intercepts the log to extract a formalized constraint.

\begin{tcolorbox}[title=HITL Pipeline Feedback Log ($b_t$), colback=red!5, colframe=red!40, fonttitle=\bfseries, breakable, before skip=4pt, after skip=4pt]
\small
\textbf{1. Real-World Feasibility Assessment:} \\
\texttt{Feasibility Score: 0 (Physically Unfeasible)} \\[4pt]
\textbf{2. Operator Colloquial Feedback (Raw Pipeline Log):} \\
\textit{"For the task of drawing number 2 with a marker, it only works on a robot like AgileX because the operator and the robot face the same direction. With Franka, we operate from the opposite side of the mechanical arm, which mirrors our controls and makes precise asymmetric tracing impossible for the data collection staff."} \\[4pt]
\textbf{3. LLM Distilled Actionable Guideline ($V$):} \\
\texttt{"Writing or precise asymmetrical character tracing fails on opposite-facing stationary arms due to spatial mapping mirror inversions; fine trajectory tracing requires co-directional mobile bases."}
\end{tcolorbox}

\begin{center} \Large $\Downarrow$ \end{center}

\noindent\textbf{Phase 3: Long-Term Memory Solidification $(\mathcal{M} \leftarrow \mathcal{M} \cup \{(K, V)\})$} \\
The framework pairs the full environmental context ($K$) with the distilled guideline ($V$) into a structural key-value node, expanding the semantic constraint boundaries of $\mathcal{M}$.

\begin{tcolorbox}[title=Memory Key-Value Node Solidification, colback=purple!5, colframe=purple!40, fonttitle=\bfseries, breakable, before skip=4pt, after skip=4pt]
\begin{lstlisting}[
    basicstyle=\ttfamily\scriptsize, 
    breaklines=true,                 
    columns=fullflexible,            
    keepspaces=true                  
]
{
  "Key_K (Full Task Context)": {
    "robot_embodiment": "dual-arm stationary robot (Franka)",
    "operator_perspective": "opposite-facing (allocentric)",
    "target_skill": "write / trace / fine asymmetric drawing",
    "target_objects": ["marker", "whiteboard"]
  },
  "Value_V (Actionable Constraint)": "Reject fine-grained asymmetric drawing or character tracing on opposite-facing embodiments without active transformation calibration; re-route precise tracing to co-directional mobile bases or substitute with coarse unidirectional manipulation primitives (e.g., pushing, wiping)."
}
\end{lstlisting}
\end{tcolorbox}

\begin{center} \Large $\Downarrow$ \end{center}

\noindent\textbf{Phase 4: Subsequent Query \& Memory-Augmented RAG Match} \\
In a subsequent generation cycle, the Proposal Generator attempts to create another fine-grained shape drawing task on the Franka robot arm. The complete contextual state triggers a high cosine similarity match against the stored Key ($K$).

\begin{tcolorbox}[title=Context Window RAG Injection, colback=orange!5, colframe=orange!40, fonttitle=\bfseries, breakable, before skip=4pt, after skip=4pt]
\small
\textbf{New Task Spec Query:} \texttt{\{Robot: Franka, Skill: draw, Task: Draw an arrow indicator\}} \\
$\longrightarrow$ \textbf{Embedding Search:} Computes $\text{CosSim}(T_{query}, K) > \theta_{\mathcal{M}}$ \\
$\longrightarrow$ \textbf{Heuristic Retrieval:} Stored Value ($V$) is actively injected into the Refiner $E_{ref}(\cdot)$ window as a hard physical boundaries constraint.
\end{tcolorbox}

\begin{center} \Large $\Downarrow$ \end{center}

\noindent\textbf{Phase 5: Refined Target Generation \& Verification ($T_{ref}$)} \\
Armed with the retrieved memory constraint, the Refiner model automatically suppresses the physical hallucination, gracefully pivoting the fine trajectory tracing into a coarse parallel stroke primitive compatible with opposite-facing control loops.

\begin{tcolorbox}[title=Final Executable Refined Output ($T_{ref}$), colback=green!5, colframe=green!40, fonttitle=\bfseries, breakable, before skip=4pt, after skip=4pt]
\begin{lstlisting}[
    basicstyle=\ttfamily\scriptsize, 
    breaklines=true,                 
    columns=fullflexible,            
    breakatwhitespace=false,         
    keepspaces=true                  
]
{
  "task_name": "robogene_franka_board_wipe",
  "task_description": "The opposite-facing robot arm utilizes a board eraser to perform coarse parallel linear wiping motions on the whiteboard surface.",
  "language_instruction": "Grasp the eraser and wipe the whiteboard surface using parallel linear strokes.",
  "objects": ["whiteboard eraser", "tabletop whiteboard"],
  "skills": ["wipe"],
  "scene_layout": {
    "whiteboard eraser": "Right Arm Workspace: Middle Right",
    "tabletop whiteboard": "Center of the table"
  },
  "task_context": "The task bypasses fine spatial inversion conflicts by employing coarse directional primitives easily manageable under opposite-facing teleoperation mapping."
}
\end{lstlisting}
\end{tcolorbox}

\begin{tcolorbox}[title=Downstream Deployment Verification, colback=green!2, colframe=green!30, breakable, before skip=2pt]
\scriptsize
\texttt{Final Feasibility Score: 1 (Successfully Executed)} \\
\textbf{Operator Verification Log:} \textit{"Coarse parallel wiping does not suffer from control mirroring difficulties. The teleoperation staff successfully completed all 250 data collection trajectories on the Franka arm without structural intervention."}
\end{tcolorbox}

\section{Evaluation Metrics and Protocol}
\label{app:metrics}

To rigorously validate the efficacy of the tasks generated by our framework, we establish a comprehensive benchmarking protocol. 
We compare \textbf{RoboGene} against four distinct baselines: a standard Rule-based method, Human expert design, and two state-of-the-art Large Foundation Models (LFMs), specifically GPT-4o~\cite{hurst2024gpt} and Gemini 2.5 Pro~\cite{comanici2025gemini}. 
To ensure statistical significance, we generated a diverse evaluation corpus consisting of 900 tasks for each method, balanced evenly across three embodiment categories: 300 single-arm, 300 dual-arm, and 300 mobile manipulation tasks. 
Given that quantifying the quality of open-ended robotic tasks is an under-explored challenge, we propose a novel suite of multi-dimensional metrics designed to assess linguistic clarity, logical coherence, asset grounding, and physical executability. 
The three judgment-based metrics are independently assessed by human
evaluators and Claude Opus 5 in Table~\ref{tab:independent_judging}, and the main text reports the average
of the two evaluations.


\textbf{Task Clarity.}
This metric evaluates the linguistic quality of the generated instruction. 
It assesses whether the task description is concise, unambiguous, and free from excessive rhetoric, ensuring interpretability by human operators during data collection. 
To mitigate subjective bias, we employ a hybrid evaluation strategy. 
We aggregate binary judgments (0 or 1) from two distinct evaluators, including Claude Opus 5 and human experts, and calculate the final score as the average of these two independent assessments.

\textbf{Robot Type Consistency.}
This metric verifies the alignment between the generated task description and the specified robot embodiment constraints $r$. 
A penalty is assigned if the task requirements contradict the robot's kinematic configuration, such as assigning a bimanual coordination task to a single-arm robot. 
Similar to task clarity, this score is derived by human evaluators and Claude Opus 5 using
the same binary rubric.

\textbf{Logical Validity.}
This metric determines if the task description adheres to real-world semantic and common-sense constraints. 
It penalizes contextually inappropriate actions, such as performing industrial machining operations within a domestic dining environment. 
This metric is assessed by human evaluators and Claude Opus 5 using
the same binary rubric.

\begin{table}[t]
    \centering
    \caption{Independent judging results.}
    \label{tab:independent_judging}

    \footnotesize
    \setlength{\tabcolsep}{4pt}
    \renewcommand{\arraystretch}{1.05}
    \setlength{\aboverulesep}{1.2pt}
    \setlength{\belowrulesep}{1.2pt}

    \begin{tabular}{@{}lcccccc@{}}
        \toprule

        \multirow{3}{*}{Method}
        & \multicolumn{3}{c}{Human Evaluator}
        & \multicolumn{3}{c}{Claude Opus 5} \\

        \cmidrule(lr){2-4}
        \cmidrule(lr){5-7}

        & Task
        & Type
        & Logical
        & Task
        & Type
        & Logical \\

        & Clarity
        & Consistency
        & Validity
        & Clarity
        & Consistency
        & Validity \\

        \midrule

        Rule-based
        & 0.5511
        & 0.3911
        & 0.4022
        & 0.5433
        & 0.4022
        & 0.4000 \\

        Human
        & 0.9622
        & 0.8378
        & 0.9544
        & 0.9567
        & 0.8311
        & 0.9578 \\

        GPT-4o
        & 0.8000
        & 0.8511
        & 0.7589
        & 0.7956
        & 0.8400
        & 0.7556 \\

        Gemini 2.5 Pro
        & 0.8900
        & 0.8711
        & 0.6700
        & 0.8878
        & 0.8811
        & 0.6689 \\

        \midrule

        \bbluecell{\textbf{RoboGene}}
        & \bbluecell{\textbf{0.9889}}
        & \bbluecell{\textbf{0.9878}}
        & \bbluecell{\textbf{0.9900}}
        & \bbluecell{\textbf{0.9933}}
        & \bbluecell{\textbf{0.9811}}
        & \bbluecell{\textbf{0.9956}} \\

        \bottomrule
    \end{tabular}
\end{table}
\textbf{Object Coverage.}
To quantify the precision of asset grounding, we define object coverage as the ratio of generated objects that exist within the valid object library $\mathcal{O}$ to the total number of unique objects proposed by the method. 
We utilize string matching algorithms to compute this metric. Formally, let $O_{gen}$ be the set of unique objects generated by a method; the coverage is calculated as $|O_{gen} \cap \mathcal{O}| / |O|$. 
A higher score indicates that the generated task set covers a broader portion of the predefined object library, which reflects stronger object-level diversity and helps mitigate long-tail object usage in VLA pre-training.


\textbf{Skill Coverage.}
Similar to object coverage, this metric measures the grounding of manipulation skills.
It is defined as the proportion of generated skills that align with the predefined skill library $\mathcal{S}$ relative to the total number of predefined skills. 
Using string matching, we calculate this as $|S_{gen} \cap \mathcal{S}| / |S|$.
A higher score indicates that the generated task set covers a broader portion of the predefined manipulation-skill library, rather than repeatedly using only frequent head skills.

\textbf{Asset Validity.} Coverage measures how broadly a method uses the predefined asset libraries, but it does not directly quantify whether generated objects or skills are valid. To explicitly measure asset-grounding accuracy and hallucination, we define object validity and skill validity as the fractions of generated object and skill types that belong to the predefined libraries: \[ \mathrm{ObjectValidity} = \frac{|O_{\mathrm{gen}} \cap O|}{|O_{\mathrm{gen}}|}, \qquad \mathrm{SkillValidity} = \frac{|S_{\mathrm{gen}} \cap S|}{|S_{\mathrm{gen}}|}. \] Higher validity indicates fewer hallucinated or unsupported assets. The corresponding results are reported in Table~\ref{tab:asset_validity}.

\textbf{Physical Feasibility.}
This is the most critical metric, evaluating the practical executability of the task in the real world. 
Unlike the semantic metrics, feasibility is assessed through physical interaction. 
We recruited human operators to attempt each task via teleoperation on the physical robot hardware. 
We report the average success rate based on five independent teleoperation trials per task. 
A trial is considered successful only if the human operator can successfully complete the objective defined in the instruction.
The task-level score is
\[
s_{\rm phys}(\tau)=\frac{1}{5}\sum_{j=1}^{5}\mathbb I[\text{trial }j\text{ succeeds}],
\]
and the reported method-level score is the average over all evaluated tasks.

\begin{table}[t]
\centering
\small
\caption{Binary rubric used for judgment-based evaluation.}
\label{tab:eval-rubric}
\begin{tabular}{p{0.20\linewidth}p{0.36\linewidth}p{0.36\linewidth}}
\toprule
Metric & Score 1 & Score 0 \\
\midrule
Task Clarity &
The task is self-contained and unambiguous; the required action, target objects,
and desired outcome can be identified without extra assumptions. &
The task is vague, underspecified, internally contradictory, or contains irrelevant
rhetoric that prevents reliable interpretation. \\

Robot Type\newline Consistency  &
All required actions are compatible with the specified robot embodiment, including
the number of arms, end-effectors, workspace, mobility, and kinematic constraints. &
The task requires unsupported capabilities, such as bimanual manipulation for a
single-arm robot, locomotion for a fixed-base robot, or manipulation outside the
reachable workspace. \\

Logical Validity &
The task is semantically plausible and contextually appropriate for the given scene. &
The task violates common-sense or contextual constraints, requests impossible object
interactions, or places actions in an inappropriate environment. \\
\bottomrule
\end{tabular}
\end{table}

\begin{tcolorbox}[
  title=Evaluation Judge Prompt,
  colback=gray!10,
  colframe=gray!50,
  fonttitle=\bfseries,
  breakable,
  before skip=6pt,
  after skip=6pt
]

\begin{lstlisting}[
  basicstyle=\ttfamily\scriptsize,
  breaklines=true,
  columns=fullflexible,
  breakatwhitespace=false,
  keepspaces=true,
  escapeinside={(*}{*)}
]
{
  "(*\textbf{system}*)":
    "You are an expert evaluator of robotic manipulation tasks. Your responsibility is to assess the quality of a generated robotic task according to three binary metrics: Task Clarity, Robot Type Consistency, and Logical Validity. You must judge only the given task specification and the provided robot/environment constraints. Do not assume additional objects, skills, robot capabilities, or scene conditions beyond the input.",

  "(*\textbf{user}*)":
    "You will be given a generated robotic task in a JSON-like format. Each task contains the following fields:

    - task_name: the name of the generated task.
    - task_description: a detailed natural-language description of the task.
    - language_instruction: the instruction that would be given to the robot or VLA model.
    - objects: the objects referenced by the task.
    - skills: the manipulation skills required by the task.
    - scene_layout: the spatial placement of objects in the robot workspace.
    - task_context: additional information about object states, poses, and scene conditions.
    - robot_type: the target robot embodiment, such as single-arm, dual-arm, or mobile manipulator.
    - robot_constraints: constraints of the target robot, including arm number, workspace, end-effector, mobility, and kinematic limitations.

    Evaluate the task using the following binary criteria.

    1. Task Clarity:
       Return 1 if the task is self-contained, specific, and unambiguous. The required action, target objects, execution order, and desired goal state should be identifiable from the task_description and language_instruction without requiring extra assumptions.
       Return 0 if the task is vague, underspecified, internally contradictory, missing a clear goal, or contains irrelevant rhetoric that prevents reliable interpretation.

    2. Robot Type Consistency:
       Return 1 if the required actions are compatible with the specified robot_type and robot_constraints. The task must respect the robot's number of arms, end-effector capabilities, reachable workspace, mobility, and kinematic limitations.
       Return 0 if the task requires unsupported capabilities, such as bimanual manipulation for a single-arm robot, locomotion for a fixed-base robot, manipulation outside the reachable workspace, or actions that the specified embodiment cannot physically perform.

    3. Logical Validity:
       Return 1 if the task is semantically plausible and contextually appropriate given the objects, skills, scene_layout, and task_context. Object states, spatial relations, actions, and outcomes must be mutually consistent.
       Return 0 if the task violates common-sense or contextual constraints, requests impossible object interactions, uses incompatible object states, or places actions in an inappropriate scene.

    Important evaluation rules:
    - Judge each metric independently.
    - Use binary labels only: 1 means the criterion is satisfied, and 0 means it is not satisfied.
    - Penalize hallucinated or unsupported assumptions.
    - Do not reward a task for being creative if it is unclear, inconsistent with the robot type, or logically invalid.
    - If the task is partially correct but has a serious ambiguity or contradiction for a metric, assign 0 for that metric.

    Input task format:
    {
      task_name: <task name>,
      task_description: <detailed task description>,
      language_instruction: <robot instruction>,
      objects: [<object 1>, <object 2>, ...],
      skills: [<skill 1>, <skill 2>, ...],
      scene_layout: {
        <object 1>: <workspace location>,
        <object 2>: <workspace location>
      },
      task_context: <object states and scene context>,
      robot_type: <target robot embodiment>,
      robot_constraints: <robot embodiment constraints>
    }

    Return only valid JSON in the following format:
    {
      \"task_clarity\": 0 or 1,
      \"robot_type_consistency\": 0 or 1,
      \"logical_validity\": 0 or 1,
      \"rationale\": {
        \"task_clarity\": \"A concise explanation for the Task Clarity score.\",
        \"robot_type_consistency\": \"A concise explanation for the Robot Type Consistency score.\",
        \"logical_validity\": \"A concise explanation for the Logical Validity score.\"
      }
    }"
}
\end{lstlisting}

\end{tcolorbox}


\begin{table}[t]
    \centering
    \caption{
        Asset validity of generated object and skill sets.
        Validity is computed as the fraction of generated object or
        skill types that belong to the predefined physical asset libraries.
    }
    \label{tab:asset_validity}

    \footnotesize
    \setlength{\tabcolsep}{4pt}
    \renewcommand{\arraystretch}{1.05}
    \setlength{\aboverulesep}{1.2pt}
    \setlength{\belowrulesep}{1.2pt}

    \begin{tabular}{@{}lcc@{}}
        \toprule

        Method
        & Object Validity
        & Skill Validity \\

        \midrule

        GPT-4o
        & 85.1\% (353/415)
        & 85.7\% (30/35) \\

        Gemini 2.5 Pro
        & 48.0\% (247/515)
        & 63.0\% (29/46) \\

        Human
        & 100.0\%
        & 100.0\% \\

        \midrule

        \bbluecell{\textbf{RoboGene}}
        & \bbluecell{\textbf{100.0\%}}
        & \bbluecell{\textbf{100.0\%}} \\

        \bottomrule
    \end{tabular}
\end{table}

\section{Implementation Details of Baselines}
\label{app:baselines}

To ensure a rigorous comparative analysis, we provide detailed implementations of the baseline methods employed in our evaluation. 
These baselines represent the spectrum of task generation approaches, ranging from rule-based and human-based algorithms to state-of-the-art foundation models operating without agentic scaffolding.

\textbf{Rule-based Method.}
This baseline utilizes a deterministic algorithm structured around a triple-nested loop. 
The system systematically iterates through the alphabetized lists of valid scenarios $\mathcal{E}$, objects $\mathcal{O}$, and skills $\mathcal{S}$ to mechanically synthesize task tuples. 
While this exhaustive combinatorial approach maximizes the theoretical coverage of the search space, it lacks semantic filtering capabilities. 
Consequently, the method produces a high volume of logically incoherent tasks, such as ``cooking a wrench,'' which are physically infeasible. 
Furthermore, the generated instructions are primarily limited to simple, single-step primitive actions, failing to capture the complexity required for generalist robot learning.

\textbf{Human-based Method.}
To evaluate manual curation, we recruited expert annotators to generate tasks, granting them access to the complete history of previously designed entries to encourage diversity in distribution. 
However, our observations identified two critical limitations in this approach. 
First, human designers exhibited significant cognitive bias, disproportionately favoring familiar settings such as Domestic and Kitchen environments while neglecting specialized domains like Industry or Medical scenarios. 
This tendency reflects an inherent difficulty for humans to conceive tasks within unfamiliar physical contexts. 
Second, the designers rapidly encountered a creativity plateau, where subsequent generations devolved into minor variations of existing tasks, such as simple object substitution or positional adjustments, rather than introducing structurally novel behaviors.

\textbf{GPT-4o.}
For the GPT-4o baseline, we leveraged the prompt engineering capabilities of the model by directly injecting the complete sets of definitions for scenarios $\mathcal{E}$, objects $\mathcal{O}$, and skills $\mathcal{S}$ into the context window. 
The model was instructed to function as an autonomous task generator, outputting tasks strictly in the required JSON format. 
Crucially, this baseline operates in an open-loop manner, lacking the feedback mechanisms, self-reflection, or external memory modules present in our framework.

\textbf{Gemini 2.5 Pro.}
Similarly, we designed a dedicated prompt for the Gemini 2.5 Pro baseline, embedding the requisite definition lists and formatting constraints akin to the GPT-4o setup. 
This baseline serves to evaluate the intrinsic raw generation capabilities of state-of-the-art foundation models when deprived of the agentic scaffolding provided by RoboGene. 
By comparing these results, we isolate the specific contributions of our diversity-driven sampling and self-reflection mechanisms.

\section{Validation of Generated Tasks}
\label{app:task_validation}
\begin{figure}[!htbp]
    \centering
    \includegraphics[
        width=0.68\linewidth,
    ]{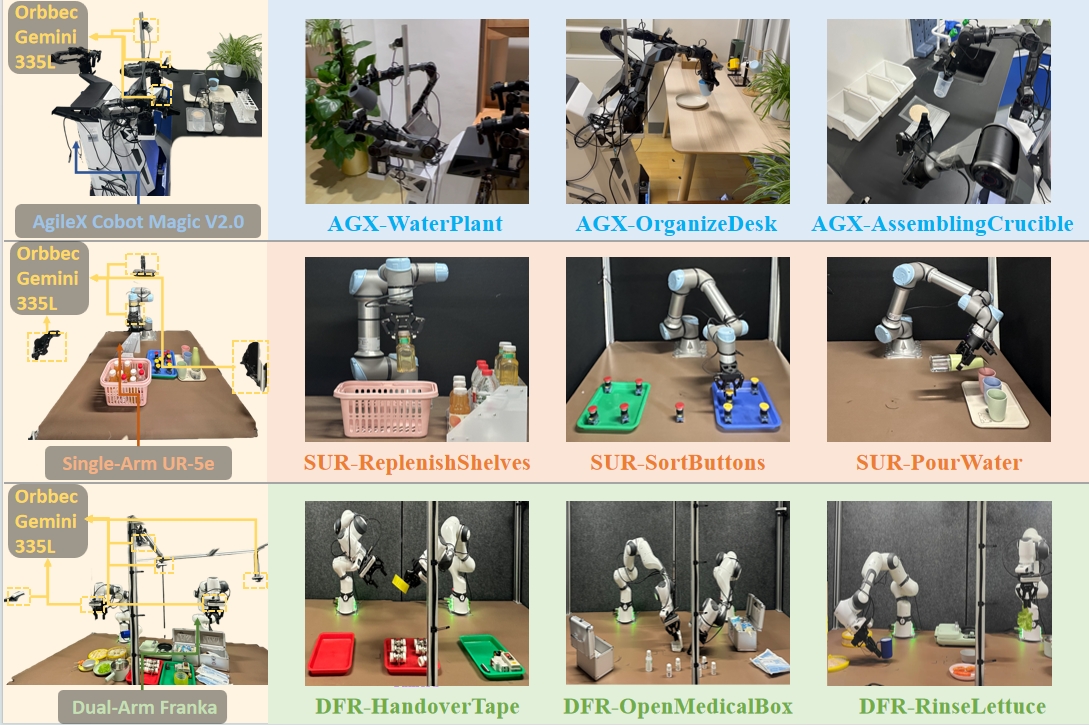}
    \vspace{-0.4em}
    \caption{Real-world experimental setup with three robots: Single-arm UR-5e (SUR), Dual-arm Franka (DFR), and AgileX Cobot Magic V2.0 (AGX).}
    \label{fig:single_experiment_setup}
    \vspace{-0.6em}
\end{figure}

To further verify the executability of RoboGene-generated tasks, we conduct representative single-task validation on three real robot embodiments: a single-arm UR-5e, a dual-arm Franka, and an AgileX mobile robot, as shown in Figure~\ref{fig:single_experiment_setup}. 
For each robot type, we generate three representative tasks using \ourmethod~and collect 250 teleoperated demonstrations per task. 
We train ACT~\cite{zhao2023act} for the stationary robots and $\pi_{0.5}$~\cite{intelligence2025pi_05} for the mobile robot, and evaluate each trained policy over 20 rollouts.
\begin{table}[t]
    \centering
    \caption{
        Validation of data collection across three robots.
        We trained on nine tasks and reported the success rates
        of various models.
    }
    \label{tab:data_validation}

    \footnotesize
    \setlength{\tabcolsep}{4pt}
    \renewcommand{\arraystretch}{1.05}
    \setlength{\aboverulesep}{1.2pt}
    \setlength{\belowrulesep}{1.2pt}

    \begin{tabular}{@{}lccc@{}}
        \toprule

        \multirow{2}{*}{Method}
        & SUR-Replenish
        & SUR-Sort
        & SUR-Pour\\
        & Shelves
        & Buttons
        & Water\\

        \midrule

        ACT
        & 60\%
        & 90\%
        & 55\% \\

        \midrule

        \multirow{2}{*}{Method}
        & DFR-Handover
        & DFR-OpenMedical
        & DFR-Rinse \\
        & Tape
        & Box
        & Lettuce \\

        \midrule

        ACT
        & 90\%
        & 85\%
        & 60\% \\

        \midrule

        \multirow{2}{*}{Method}
        & AGX-Water
        & AGX-Organize
        & AGX-Assembling \\
        & Plant
        & Desk
        & Crucible \\

        \midrule

        $\pi_{0.5}$
        & 45\%
        & 60\%
        & 70\% \\

        \bottomrule
    \end{tabular}
\end{table}
As shown in Table~\ref{tab:data_validation}, the generated tasks can be successfully executed across all three embodiments, with task-level success rates ranging from 45\% to 90\%. 
These results indicate that RoboGene produces tasks that are physically deployable in real-world data collection while retaining different levels of manipulation difficulty across fixed-base and mobile settings.

\section{Ablation Variant Definitions}
\label{app:ablation_details}

We provide implementation details for the ablation variants reported
in Table~\ref{tab:component_ablation}.

\paragraph{w/o LFU.}
We remove the Least Frequently Used sampling strategy while retaining
Self-Reflection and long-term memory.

\paragraph{w/o Self-Reflection.}
We retain LFU sampling and long-term memory but remove the iterative
critic--refiner loop, resulting in single-pass constrained generation.

\paragraph{w/o Memory.}
We retain LFU sampling and Self-Reflection but disable the retrieval
and consolidation of human-verified execution feedback.

\paragraph{BRS+LLM.}
Balanced Random Sampling with LLM generation samples candidate
objects and skills from the predefined libraries without LFU,
iterative Self-Reflection, or long-term memory.

\section{Additional Experiments}
\label{sec:add_exp}
\subsection{Downstream Fine-Tuning with 50 Demonstrations}
\label{app:fifty_demo_finetuning}
To further evaluate the efficacy of the data generated by our framework, we conducted additional experiments focusing on the downstream transfer capabilities of Vision-Language-Action (VLA) models. 
Specifically, we investigated whether pre-training on diverse, synthetically generated tasks facilitates better adaptation to unseen real-world scenarios compared to baselines. 
We constructed distinct pre-training datasets, each comprising 150 single-arm and 150 dual-arm tasks generated by Human experts, GPT-4o, Gemini 2.5 Pro, and RoboGene, respectively.

We employed the $\pi_{0}$~\cite{black2025pi0} as the backbone policy for these experiments. 
Following the pre-training phase on each respective dataset, we fine-tuned the models on a set of five unseen dual-arm tasks: DFR-GrillSkewers, DFR-HangCups, DFR-BuildBlocks, DFR-LubricateGears, and DFR-WeighBeaker. 
To further examine whether the benefit of RoboGene pre-training persists under stronger task-specific supervision, we conduct an additional fine-tuning evaluation with 50 demonstrations per downstream task. We report the average success rate over 20 real-world rollouts for each task.

The quantitative results are summarized in Table~\ref{tab:add_pretrain}. 
The policy pre-trained on the RoboGene-generated tasks consistently outperforms the baselines, achieving the highest average success rate of 48\%. 
Notably, RoboGene demonstrates a significant advantage in complex tasks such as DFR-BuildBlocks, where it achieves a 30\% success rate compared to 5\% for Human-curated data and 15\% for other Foundation Models. 

\begin{table*}[t]
    \centering
    \caption{
        Additional comparative results of downstream fine-tuning
        performance on unseen dual-arm tasks.
        We report the success rates (\%) of the $\pi_{0}$ model
        pre-trained on different datasets and fine-tuned with
        50 demonstrations per task.
    }
    \label{tab:add_pretrain}

    \footnotesize
    \setlength{\tabcolsep}{4pt}
    \renewcommand{\arraystretch}{1.05}
    \setlength{\aboverulesep}{1.2pt}
    \setlength{\belowrulesep}{1.2pt}

    \begin{tabular}{@{}lcccccc@{}}
        \toprule

        \multirow{2}{*}{Method}
        & DFR-Grill
        & DFR-Hang
        & DFR-Build
        & DFR-Lubricate
        & DFR-Weigh
        & \multirow{2}{*}{Average} \\

        & Skewers
        & Cups
        & Blocks
        & Gears
        & Beaker
        & \\

        \midrule

        Human
        & 75
        & 40
        & 5
        & 45
        & 25
        & 38 \\

        GPT-4o
        & 55
        & 40
        & 15
        & 50
        & 20
        & 36 \\

        Gemini 2.5 Pro
        & 80
        & 30
        & 15
        & 35
        & 15
        & 35 \\

        \midrule
        \bbluecell{\textbf{RoboGene}}
        & \bbluecell{\textbf{80}}
        & \bbluecell{\textbf{50}}
        & \bbluecell{\textbf{30}}
        & \bbluecell{\textbf{55}}
        & \bbluecell{\textbf{25}}
        & \bbluecell{\textbf{48}} \\

        \bottomrule
    \end{tabular}
\end{table*}

This performance gap highlights that while human data is physically valid, it often lacks the structural diversity required to generalize to complex multi-stage manipulations. 
Furthermore, while GPT-4o and Gemini 2.5 Pro provide some level of diversity, their lack of physical grounding often leads to suboptimal pre-training priors. 
In contrast, RoboGene successfully balances diversity with physical feasibility, enabling the VLA model to learn robust representations that transfer effectively to novel, unseen tasks.

\subsection{Broader OOD Generalization}
\label{app:ood_ci}

We conduct a broader out-of-distribution (OOD) evaluation on five
held-out dual-arm manipulation tasks: DFR-GrillSkewers,
DFR-HangCups, DFR-BuildBlocks, DFR-LubricateGears, and
DFR-WeighBeaker. For this experiment, each policy is first
pre-trained on the dataset generated by the corresponding task
generation method and then fine-tuned with 50 demonstrations per
downstream task.

We evaluate each fine-tuned policy under five OOD shifts:
novel objects, background variations, static distractors,
illumination changes, and instruction changes. For each
task--shift combination, we conduct three independent runs of
20 real-world trials, resulting in 60 trials per shift and
300 trials per downstream task. Therefore, each method is evaluated
over 1,500 real-world OOD trials in total.










\begin{table}[h]
    \centering
    \vspace{-1em}

    \caption{OOD generalization SR(\%) with 95\% Wilson CI.}
    \label{tab:OOD_shifts}

    \scriptsize
    \setlength{\tabcolsep}{1.6pt}
    \renewcommand{\arraystretch}{1.05}

    \resizebox{\columnwidth}{!}{%
    \begin{tabular}{@{}lcccccc@{}}
        \toprule

        \multicolumn{1}{c}{Method}
        & Grill
        & Hang
        & Build
        & Lubr.
        & Weigh
        & Avg. \\

        \midrule

        Human
        & 42.0 [36.6,47.7]
        & 15.0 [11.4,19.5]
        & 2.0 [0.9,4.3]
        & 24.0 [19.5,29.1]
        & 12.0 [8.8,16.2]
        & 19.0 [17.1,21.1] \\

        GPT-4o
        & 30.0 [25.1,35.4]
        & 8.0 [5.4,11.6]
        & 6.0 [3.8,9.3]
        & 26.0 [21.4,31.2]
        & 9.0 [6.3,12.8]
        & 15.8 [14.0,17.7] \\

        Gemini 2.5 Pro
        & 46.0 [40.4,51.7]
        & 12.0 [8.8,16.2]
        & 7.0 [4.6,10.5]
        & 17.0 [13.2,21.7]
        & 7.0 [4.6,10.5]
        & 17.8 [15.9,19.8] \\

        \bbluecell{\textbf{RoboGene}}
        & \bbluecell{\textbf{67.0 [61.5,72.1]}}
        & \bbluecell{\textbf{40.0 [34.6,45.6]}}
        & \bbluecell{\textbf{25.0 [20.4,30.2]}}
        & \bbluecell{\textbf{48.0 [42.4,53.6]}}
        & \bbluecell{\textbf{21.0 [16.8,26.0]}}
        & \bbluecell{\textbf{40.2 [37.7,42.7]}} \\

        \bottomrule
    \end{tabular}%
    }

    \vspace{-2em}
\end{table}
The task-level success rate is computed by pooling the 300 trials
across the five shifts. The average success rate is computed by
pooling all 1,500 trials across the five tasks. We report 95\%
Wilson confidence intervals to account for the finite number of
real-world trials. Table~\ref{tab:OOD_shifts} presents the complete
results.
\subsection{Results on Task Semantic Diversity}
\label{app:diverse}

Finally, we quantify task description diversity using BLEU~\cite{papineni2002bleu}, ROUGE-L~\cite{lin2004rouge}, and cosine similarity metrics in Table~\ref{table:semantic_diversity_results}. 
Lower scores in this context denote less lexical overlap and higher diversity. 
\ourmethod~consistently records the lowest similarity scores, such as a BLEU-4 score of 1.75 compared to 2.71 for GPT-4o, indicating that our tasks are described using distinct phrasing and structures.
In contrast, rule-based and human-designed tasks exhibit high repetition (BLEU-1 $>$ 70), reflecting a lack of task variety. 
This confirms that \ourmethod~successfully leverages the linguistic generative power of LFMs to produce diverse instructions, effectively mitigating mode collapse.


\begin{table}[!t]
\centering
\caption{Results on task semantic diversity. Lower scores indicate lower similarity between task and greater dataset diversity.}
\vspace{2pt}
\label{table:semantic_diversity_results}

\setlength{\tabcolsep}{6pt}

\renewcommand{\arraystretch}{0.88}   
\setlength{\aboverulesep}{2pt}       
\setlength{\belowrulesep}{2pt}       
\setlength{\extrarowheight}{0pt}

\begin{tabular}{c|ccccc}
\toprule
\textbf{Metric / Method} & \bbluecell{\textbf{RoboGene}} & GPT-4o & Gemini 2.5 Pro & Rule-based & Human \\
\midrule
BLEU-1 ($\downarrow$) & \bbluecell{\textbf{28.93}} & 36.61 & 38.97 & 80.32 & 72.80 \\
BLEU-2 ($\downarrow$) & \bbluecell{\textbf{6.62}} & 9.84 & 9.76 & 67.82 & 65.62 \\
BLEU-3 ($\downarrow$) & \bbluecell{\textbf{3.15}} & 4.82 & 5.73 & 62.11 & 60.19 \\
BLEU-4 ($\downarrow$) & \bbluecell{\textbf{1.75}} & 2.71 & 3.23 & 56.45 & 55.55 \\
ROUGE-L ($\downarrow$) & \bbluecell{\textbf{0.1918}} & 0.2512 & 0.2767 & 0.5786 & 0.4886 \\
Cosine Similarity ($\downarrow$) & \bbluecell{\textbf{29.07}} & 34.64 & 34.63 & 69.84 & 68.88 \\
\bottomrule
\end{tabular}
\vspace{-15pt}
\end{table}

\subsection{Human Effort and Memory Analysis}
\label{sec:human_effort}

\paragraph{Human effort.}
We estimate human effort for generating and validating 900 tasks.
Manual task design takes about 10.9 minutes per task, while RoboGene automates this stage.
Each task requires verification ($\sim$10 s) and five teleoperation trials ($\sim$30 s/trial);
failed tasks are revised ($\sim$60 s/correction), with an additional $\sim$30 s/correction for memory summarization in RoboGene.
Using correction rates of 9.78\% and 2.01\%, the estimated total effort decreases from 204.97 to 40.45 hours, an 80.3\% reduction (Table~\ref{tab:human_effort}).

\begin{figure}[H]
    \centering
    \includegraphics[
        width=0.55\linewidth,
        trim=1 1 1 1,
        clip
    ]{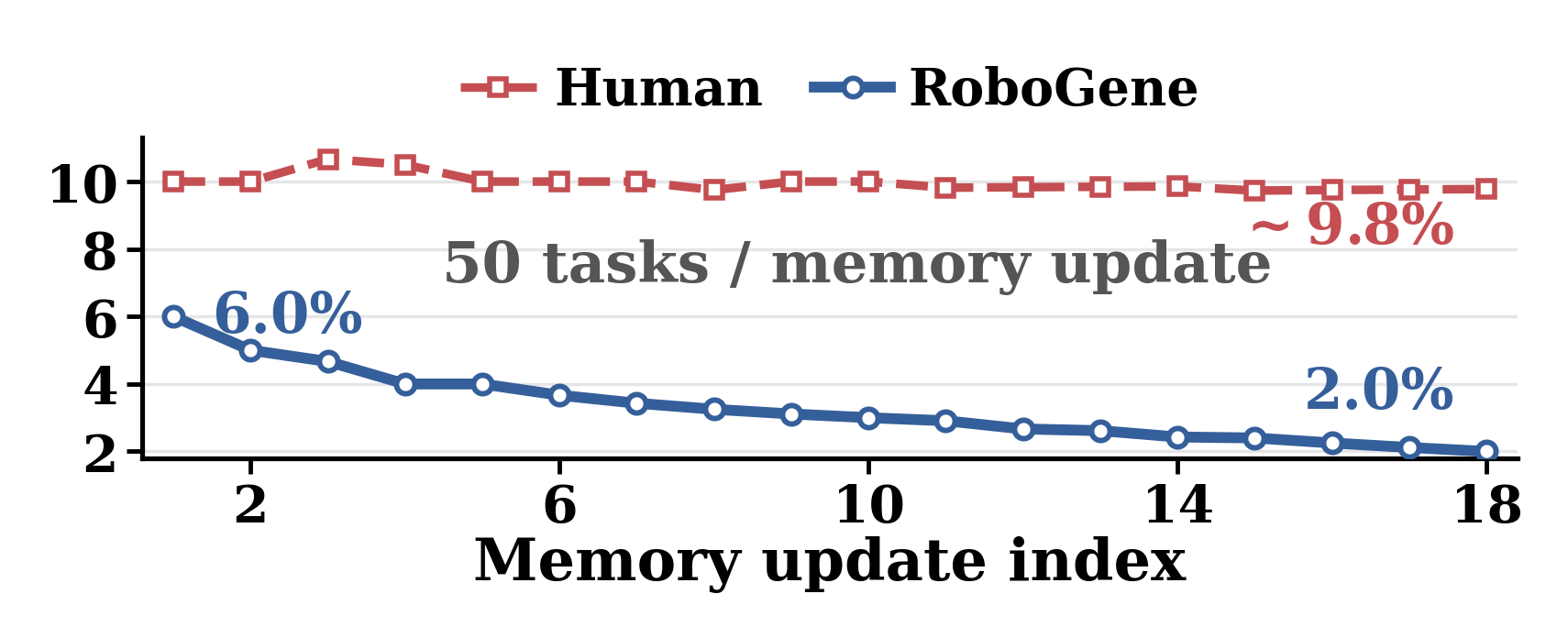}

    \vspace{-0.5em}
    \caption{
        Correction rate as human-verified feedback accumulates in long-term memory.
    }
    \label{fig:correction_rate}

    \vspace{-0.8em}
\end{figure}

\begin{table}[t]
    \centering
    \caption{
        Estimated human effort for generating and validating 900 tasks.
        Time is reported in hours except for the last column.
    }
    \label{tab:human_effort}

    \setlength{\tabcolsep}{6pt}
    \renewcommand{\arraystretch}{0.88}
    \setlength{\aboverulesep}{2pt}
    \setlength{\belowrulesep}{2pt}
    \setlength{\extrarowheight}{0pt}

    \begin{tabular}{lccccccc}
        \toprule
        Method
        & Design
        & Verify
        & Execution
        & Corr.
        & Memory
        & Total (h)
        & min/task \\
        \midrule

        Human
        & 163.50
        & 2.50
        & 37.50
        & 1.47
        & --
        & 204.97
        & 13.66 \\

        \bbluecell{\textbf{RoboGene}}
        & \bbluecell{\textbf{--}}
        & \bbluecell{\textbf{2.50}}
        & \bbluecell{\textbf{37.50}}
        & \bbluecell{\textbf{0.30}}
        & \bbluecell{\textbf{0.15}}
        & \bbluecell{\textbf{40.45}}
        & \bbluecell{\textbf{2.70}} \\

        \midrule
        \multicolumn{6}{r}{\textbf{Human-effort reduction}}
        & \multicolumn{2}{c}{\bbluecell{\textbf{80.3\%}}} \\
        \bottomrule
    \end{tabular}

    \vspace{-15pt}
\end{table}
\paragraph{Memory accumulation.}
As human-verified feedback accumulates, RoboGene's correction rate decreases from about 6.0\% to 2.0\%, while the human baseline remains near 10\% (Figure~\ref{fig:correction_rate}).
This indicates that long-term memory progressively reduces the need for corrective intervention.



\section{Prompt Structure for Agentic Nodes}
\label{app:prompts}

The efficacy of RoboGene relies on carefully engineered prompts that guide the Large Foundation Models (LFMs) within each node of the framework. 
Below, we outline the structural composition of the prompts used for each node.

\textbf{Proposal Generator Prompt.}
The generator prompt is designed to synthesize the initial task draft $T_{raw}$. 
It integrates the diversity constraints derived from the sampling module. The prompt structure includes:
(1) \textit{Role Definition:} You are an expert robotic task designer specializing in [Robot Type] manipulation.
(2) \textit{Constraint Injection:} You must generate a task within the [Scenario Name] environment. 
You are restricted to using objects from this candidate list: [List $O_t$] and skills from: [List $S_t$].
(3) \textit{Format Specification:} Output the task strictly in the defined JSON schema, ensuring all fields such as spatial layout and linguistic instructions are populated.

\begin{tcolorbox}[
  title=Proposal Generator Prompt,
  colback=gray!10,
  colframe=gray!50,
  fonttitle=\bfseries,
  breakable,
  before skip=6pt,
  after skip=6pt
]

\begin{lstlisting}[
    basicstyle=\ttfamily\scriptsize,
    breaklines=true,
    columns=fullflexible,
    breakatwhitespace=false,
    keepspaces=true,
    escapeinside={(*}{*)}
]
{
  "(*\textbf{system}*)":
    "You are an expert in bimanual robotic manipulation and task design. Your role is to generate novel,embodied plausibility, and diverse manipulation tasks for tabletop environments. You must ensure all tasks are kinematically plausible, stable, and involve meaningful bimanual coordination.",

  "(*\textbf{user}*)":
    "Task: Design diverse dual-arm manipulation scenarios based on the following specifications:

     1. Operational Constraints and Resources:

       (1) Output format requirements in JSON Structure:
           Each task must be a JSON object containing:
        - 'task_name': Following the pattern 'RoboGene_robot_action'.
        - 'task_description': A detailed description.
        - 'language_instruction': A precise English command passed into the VLA model.
        - 'objects'
        - 'skills'
        - 'scene_layout': The tabletop is divided into two symmetrical 3x3 grids corresponding to the left and right arm workspaces. The indexing follows a row-major order.
        - "task_context": explains how the object is positioned and its physical state.

     2. Permissible Assets:
     
        Only items listed in 'asset.csv' and actions defined in 'Skills.txt' are allowed.

     3. Reference Example:

        {
          "task_name": "robogene_dual_pick_and_place_white_black_cup",
          "task_description": "Both robotic arms are used to manipulate two cups of different colors. A white cup and a black cup are initially placed in the upper-right area of the table. The robot visually calibrates the objects, grasps the cups, and places them sequentially on the table in a specified order, ensuring a rim-to-rim distance of 15 centimeters between the two cups.",
          "language_instruction": "Using both arms, pick up the white cup and the black cup from the upper-right area of the table and place them on the table in the specified order, keeping a 15 cm distance between the rims of the two cups.",
          "objects": [
            "White cup",
            "Black cup"
          ],
          "skills": [
            "pick_and_place"
          ],
          "scene_layout": {
            "White cup": "Right Arm Workspace: Upper Right",
            "Black cup": "Right Arm Workspace: Upper Right"
          },
          "task_context": "Both cups are upright and stable on the tabletop. They are initially placed close to each other in the upper-right area. The table surface is flat and unobstructed, allowing sequential grasping and precise placement with a fixed rim-to-rim distance."
        }
    

     4. Safety and Feasibility Constraints:

       (1) Avoid trivial tasks:
           Avoid creating tasks that can be completed with just one arm.
       (2) Moderated Force:
           Contact-based actions (e.g., pressing) must use moderated force to ensure stability and avoid hardware damage.
       (3) Collision Avoidance:
           Strictly avoid high-impact trajectories.

     5. Final Checklist:

       - Asset Consistency:
           Strictly adhere to provided asset and skill taxonomies.
       - Logical Flow:
           Tasks must be physically executable and logically sequenced for dual-arm coordination.
       - Diversity:
           Maximize the variety of object-skill combinations."
}
\end{lstlisting}

\end{tcolorbox}

\textbf{Task Novelty Evaluator Prompt}
This evaluator ensures the generated task contributes to the dataset's diversity. The prompt instructs the model to:
(1) \textit{Analyze Complexity:} Assess the proposed task for interaction richness (e.g., tool use, deformability) versus simple pick-and-place primitives.
(2) \textit{Check Redundancy:} Compare the task logic against common patterns to identify trivial or repetitive designs.
(3) \textit{Output Decision:} Return a binary decision (Yes / No) and a critique describing the novelty level.
\begin{tcolorbox}[
  title=Task Novelty Evaluator Prompt,
  colback=gray!10,
  colframe=gray!50,
  fonttitle=\bfseries,
  breakable,
  before skip=6pt,
  after skip=6pt
]

\begin{lstlisting}[
    basicstyle=\ttfamily\scriptsize,
    breaklines=true,
    columns=fullflexible,
    keepspaces=true,
    escapeinside={(*}{*)}
]
{
  "(*\textbf{system}*)": "You are an expert evaluator for robotic manipulation task datasets. Your role is to analyze task distributions, identify potential imbalances in skill usage, and assess whether a given set of tasks is overly biased toward common manipulation paradigms or lacks novelty and rare skill types. Your analysis should be objective, concise, and suitable for inclusion in an academic context.",

  "(*\textbf{user}*)": "You will be provided with structured task descriptions. Based on the given information, analyze whether there exists a task distribution imbalance, with particular attention to skill diversity, novelty, and paradigm bias.

  1. Task Analysis Procedure:

  (1)Task Comprehension:
     Carefully examine each task, including the task name, task description, language instruction, skill name and so on.
  (2)Skill Distribution Analysis:
     Count the frequency of each skill across the task set. Evaluate whether skill usage is balanced or dominated by a small number of common skills. Consider whether additional, less frequent skills could be introduced to improve overall balance.
  (3)Novelty and Rarity Assessment:
     Determine whether the tasks involve novel or rare skill types, or whether they rely exclusively on well-established manipulation skills without introducing new operational challenges.
  (4)Paradigm Bias Evaluation:
     Analyze whether the tasks are biased toward conventional manipulation paradigms (e.g., repetitive pick-and-place patterns), thereby lacking diversity or uniqueness.
  (5)Final Judgment and Explanation:
     If a task distribution imbalance is identified, mark the result as "No". If no imbalance is found, mark the result as "Yes". Provide a brief explanation supporting your judgment, including key observations and reasoning.

  2. Input Data Structure:

     The input will be provided in a JSON-like format and may include:
        - 'task_name': Following the pattern 'RoboGene_robot_action'.
        - 'task_description': A detailed description.
        - 'language_instruction': A precise English command passed into the VLA model.
        - 'objects'
        - 'skills'
        - 'scene_layout': The tabletop is divided into two symmetrical 3x3 grids corresponding to the left and right arm workspaces. The indexing follows a row-major order.
        - "task_context": explains how the object is positioned and its physical state.

  3. Output Format:

     After completing your analysis, produce a response strictly in the following format:

     - Feasibility: [Yes / No]
     - Analysis: [A concise justification grounded in the evaluation dimensions above]

  4. Examples:

     Example 1
     Input:
    {
      "task_name": "RoboGene_dual_Franka_pick_and_place_white_and_black_cup",
      "task_description": "Both robotic arms are used to manipulate two cups of different colors. A white cup and a black cup are initially placed in the upper-right area of the table. The robot visually calibrates the objects, grasps the cups using coordinated dual-arm manipulation, and places them sequentially on the table in a specified order. The final placement must ensure a rim-to-rim distance of 15 centimeters between the two cups.",
      "language_instruction": "Use both arms to pick up the white cup and the black cup from the upper-right area of the table and place them on the table in the specified order, ensuring a 15 cm distance between the rims of the two cups.",
      "objects": [
        "White cup",
        "Black cup"
      ],
      "skills": [
        "pick_and_place"
      ],
      "scene_layout": {
        "White cup": "Right Arm Workspace: Upper Right",
        "Black cup": "Right Arm Workspace: Upper Right"
      },
      "task_context": "Both cups are upright and stable on a flat tabletop surface. They are initially positioned close to each other in the upper-right area. No external obstacles are present. During placement, the cups must remain upright, and the rim-to-rim distance between the white cup and the black cup must be maintained at 15 centimeters."
    }
     Output:
     - Feasibility: No
     - Analysis: The task set is overly dominated by the pick_and_place skill, leading to a pronounced imbalance in skill utilization. More advanced manipulation skills, including assembly, rotation, and force control, are entirely absent. This lack of skill diversity limits the representational richness and execution difficulty of the task set, and incorporating non-pick-and-place skills would be essential for achieving a more balanced and realistic evaluation."
}
\end{lstlisting}

\end{tcolorbox}

\textbf{Constraint Adherence Evaluator Prompt}
This node enforces the strict boundaries set by the sampling strategy. The prompt requires the model to:
(1) \textit{Verify Entities:} Confirm that all objects and skills mentioned in the task description exist strictly within the provided candidate sets $O_t$ and $S_t$.
(2) \textit{Check Hallucination:} Identify any invented objects or physical properties not present in the input metadata.
(3) \textit{Consistency Check:} Ensure the task description logically aligns with the selected scenario category.

\begin{tcolorbox}[
  title=Constraint Adherence Evaluator Prompt,
  colback=gray!10,
  colframe=gray!50,
  fonttitle=\bfseries,
  breakable,
  before skip=6pt,
  after skip=6pt
]

\begin{lstlisting}[
    basicstyle=\ttfamily\scriptsize,
    breaklines=true,
    columns=fullflexible,
    breakatwhitespace=false,
    keepspaces=true,
    escapeinside={(*}{*)}   
]
{
  "(*\textbf{system}*)":
    "You are an expert task consistency reviewer for structured dual-arm robotic manipulation tasks.
     Your role is to critically evaluate, validate, and refine task definitions by strictly verifying their alignment with provided object and skill inventories.
     You must operate as a deterministic critic, not a creative generator.
     You must:
       - Rely only on the provided reference lists.
       - Avoid assumptions, commonsense inference, or synonym expansion.
       - Apply exact string matching when validating objects and skills.
       - Produce clear, actionable, and academically rigorous feedback suitable for downstream automated correction or benchmarking.",

  "(*\textbf{user}*)":
    "You will be provided with structured task descriptions. Based on the given information, follow the requirements below.

    1. Consistency Verification

       (1) Skill Validation:
           - Verify whether the skill name exactly matches an entry in the skill list.
       (2) Object Validation:
           - Verify whether the object name exactly matches an entry in the object asset list.
       (3) JSON Validation:
           - Check whether the JSON-like format is strictly followed.
       Validation Rules:
         - Any object or skill not found verbatim in the reference lists must be marked as invalid.
         - Do not assume common objects, aliases, translations, or semantic equivalents are acceptable.
         - Exact string-level correspondence is mandatory.

    2. Input Data Structure:

     The input will be provided in a JSON-like format and may include:
        - 'task_name': Following the pattern 'RoboGene_robot_action'.
        - 'task_description': A detailed description.
        - 'language_instruction': A precise English command passed into the VLA model.
        - 'objects'
        - 'skills'
        - 'scene_layout': The tabletop is divided into two symmetrical 3x3 grids corresponding to the left and right arm workspaces. The indexing follows a row-major order.
        - "task_context": explains how the object is positioned and its physical state.

    3. Output Format:

     After completing your analysis, produce a response strictly in the following format:
     
     - Feasibility: [Yes / No]
     - Analysis: [A concise justification grounded in the evaluation dimensions above]


    4. Examples:
    
       Example 1  
       Input:
        {
          "task_name":"RoboGene_dual_Franka_rotate_and_place_blackboard",
          "task_description": "The robot manipulates a wooden blackboard by rotating it 90 degrees from its initial orientation and placing it at the center of the table. After rotation and placement, the long side of the board must be aligned parallel to the table edge.",
          "language_instruction": "Rotate the wooden board by 90 degrees and place it at the center of the table with its long side parallel to the table edge.",
          "objects": [
            "Blackboard"
          ],
          "skills": [
            "rotate",
            "place"
          ],
          "scene_layout": {
            "Blackboard": "Center of the table"
          },
          "task_context": "The blackboard is initially placed flat on the table with its long side perpendicular to the table edge. The board is stable, unobstructed, and graspable. During manipulation, the board must remain in contact with the tabletop, and the final placement requires the board's long edge to be parallel to the table edge."
        }
       Output:
         - Feasibility: Yes
         - Analysis: Both the skill name and the object name exactly match entries in the corresponding reference lists. No inconsistencies were detected.
    
       Example 2  
       Input:
        {
          "task_name":"RoboGene_dual_Franka_rotate_and_place_wooden_board_correctly",
          "task_description": "The robot manipulates a wooden board by grasping it, rotating it by 90 degrees from its initial orientation, and placing it at the center of the table. After placement, the long side of the wooden board must be aligned parallel to the table edge.",
          "language_instruction": "Rotate the wooden board by 90 degrees and place it at the center of the table, ensuring that its long side is parallel to the table edge.",
          "objects": [
            "Wooden board"
          ],
          "skills": [
            "Rotate and Place"
          ],
          "scene_layout": {
            "Wooden board": "Center of the table"
          },
          "task_context": "The wooden board is initially placed flat on the tabletop with its long side perpendicular to the table edge. The board is stable, unobstructed, and fully graspable. The task requires controlled rotation followed by precise placement so that the final orientation has the long side parallel to the table edge."
        }
       Output:
         - Feasibility: No
         - Analysis: The skill name "Rotate and Place" does not exist in the provided skill list. Suggested replacements include: "rotate" from the reference list.
           The object name "Wooden Board" is invalid and does not appear in the object asset list. Suggested replacements include: "blackboard" from the reference list."
}
\end{lstlisting}

\end{tcolorbox}
\textbf{Embodied Plausibility Evaluator Prompt}
This node focuses on qualitative task feasibility. The prompt requires the model to:
(1) \textit{Embodiment Check:} Check whether the generated task is consistent with the specified robot embodiment.
(2) \textit{Feasibility Check:} Verify consistency with object affordances, action order, and commonsense physical constraints.
(3) \textit{Revision Check:} Identify clear infeasible or underspecified cases and provide actionable revision suggestions.
\begin{tcolorbox}[
  title=Embodied Plausibility Evaluator Prompt,
  colback=gray!10,
  colframe=gray!50,
  fonttitle=\bfseries,
  breakable,
  before skip=6pt,
  after skip=6pt
]

\begin{lstlisting}[
    basicstyle=\ttfamily\scriptsize,
    breaklines=true,
    columns=fullflexible,
    breakatwhitespace=false,
    keepspaces=true,
    escapeinside={(*}{*)}
]
{
  "(*\textbf{system}*)":
    "You are a qualitative Robotic Task Feasibility Critic. Your role is to assess whether a structured robotic task proposal is plausibly executable under the specified robot embodiment, object properties, scene layout, and task sequence. Focus on identifying clear infeasible cases, ambiguous task descriptions, and actionable ways to revise the task.",

  "(*\textbf{user}*)":
    "You will be provided with a structured task description in a JSON-like format. Evaluate whether the task is qualitatively feasible for the specified robot embodiment.

     1. Evaluation Criteria:

       (1) Robot Capability and Embodiment Consistency:
           Check whether the task matches the specified robot type and manipulation capability.
           Flag clear embodiment mismatches, such as:
             - asking a single-arm robot to unscrew a bottle cap without a fixture or stabilizing mechanism;
             - assigning simultaneous two-arm coordination to a single-arm robot;
             - requiring one gripper to hold or manipulate two distant objects at the same time;
             - asking a mobile robot to manipulate an object before reaching the object's location.

       (2) Object-Skill Affordance:
           Check whether each skill is compatible with the object properties.
           Flag clear object-skill mismatches, such as:
             - folding a rigid object;
             - pouring from a non-container object;
             - opening an object with no openable part;
             - inserting an object into another object with no hole, slot, or receptacle;
             - wiping, grasping, or pushing an object without a plausible contact mode.

       (3) Action-Order and State Consistency:
           Check whether the task sequence follows a coherent order and maintains consistent object states.
           Flag inconsistent task decompositions, such as:
             - placing an object before it has been picked up;
             - manipulating an object before navigating to or reaching it;
             - putting an item into a closed drawer or cabinet without first opening it;
             - requiring the same object to be in two different locations or states at the same time;
             - skipping a necessary stabilization, grasping, opening, or placement step.

       (4) Dual-Arm Interaction Consistency:
           For dual-arm tasks, check whether the two arms are assigned spatially and functionally coherent roles.
           If one arm stabilizes an object while the other manipulates it, the object should be located in a region accessible to both arms, and the two arms should have complementary roles.
           Flag problematic bimanual tasks, such as:
             - both arms pulling the same object in conflicting directions;
             - one arm grasping a cup on one side of the table while the other opens a drawer on the other side, but the task describes them as performing a tightly coupled insertion;
             - requiring precise two-arm cooperation on objects that are described as spatially unrelated;
             - omitting which arm stabilizes and which arm performs the manipulation when the task requires two contacts.

       (5) Mobile Manipulation Consistency:
           For mobile manipulation tasks, check high-level navigation-manipulation consistency.
           Flag inconsistent mobile tasks, such as:
             - navigating to the kitchen but picking up an object described as being in the office;
             - picking up an object before navigating to its location;
             - delivering an object to an unspecified destination;
             - performing manipulation in a scene where the required target object or required affordance is absent;
             - placing an object into a closed container or cabinet without an opening step.

       (6) Commonsense Physical Plausibility:
           Check for obvious violations of everyday physical plausibility.
           Flag cases such as:
             - unsupported objects floating in the air;
             - stacking objects with clearly unstable contact, such as balancing a marker on a pencil;
             - manipulating an object without contact;
             - requiring a rigid object to deform;
             - relying on an object to remain stable despite an obviously insufficient support surface.

       (7) Revision Suggestions:
           If the task is infeasible or ambiguous, provide concise suggestions to make it more feasible.
           Useful suggestions include:
             - add a stabilizing step or fixture;
             - clarify left-arm and right-arm roles;
             - move the object into an appropriate workspace;
             - specify the target object's location;
             - add a navigation step before manipulation;
             - specify the delivery destination;
             - replace an incompatible object-skill pair;
             - decompose an overly coupled operation into sequential sub-tasks.

     2. Input Data Structure:
         The input will be provided in a JSON-like format and may include:
           - robot type or embodiment;
           - task name;
           - natural-language instruction;
           - sub-task sequence;
           - skill identifiers;
           - object list and object attributes;
           - scene layout or object locations;
           - navigation target or destination;
           - task context.

     3. Output Format:
         After completing your analysis, produce a response strictly in the following format:

         - Feasibility: [Yes / No]
         - Analysis: [Concise justification grounded in the criteria above. If the task is infeasible or ambiguous, include concrete revision suggestions.]

     4. Examples:

       Example 1
       Input:
        {
          "robot_type": "single-arm robot",
          "task_name": "single_arm_unscrew_bottle_cap",
          "task_description": "The robot grasps a bottle cap and unscrews it from a bottle on the table.",
          "objects": ["bottle", "bottle cap"],
          "skills": ["grasp", "twist", "unscrew"],
          "scene_layout": {
            "bottle": "center of table"
          },
          "task_context": "The bottle is not fixed by any fixture, and no stabilizing contact is specified."
        }
       Output:
         - Feasibility: No
         - Analysis: The task is underspecified for a single-arm robot because unscrewing typically requires stabilizing the bottle while twisting the cap. Add a fixture, a stabilizing contact, or use a dual-arm setup with one arm holding the bottle and the other twisting the cap.

       Example 2
       Input:
        {
          "robot_type": "dual-arm robot",
          "task_name": "dual_arm_cup_drawer_insertion",
          "task_description": "The left arm grasps a cup on the left side of the table while the right arm opens a drawer on the right side, and both arms synchronously insert the cup into the drawer.",
          "objects": ["cup", "drawer"],
          "skills": ["grasp", "open", "insert"],
          "scene_layout": {
            "cup": "far left side of table",
            "drawer": "far right side of table"
          },
          "task_context": "The two objects are spatially separated, but the task requires tightly coupled insertion."
        }
       Output:
         - Feasibility: No
         - Analysis: The task assigns a tightly coupled bimanual insertion to two spatially separated interactions. Move the cup and drawer into a common interaction region, or decompose the task into sequential steps: grasp the cup, move it near the drawer, open the drawer, and then insert the cup.

       Example 3
       Input:
        {
          "robot_type": "mobile manipulator",
          "task_name": "mobile_pick_book_from_office",
          "task_description": "The robot navigates to the kitchen and picks up a book from the office.",
          "objects": ["book"],
          "skills": ["navigate", "pick"],
          "scene_layout": {
            "book": "office"
          },
          "navigation_goal": "kitchen",
          "task_context": "The target object is not located at the specified navigation goal."
        }
       Output:
         - Feasibility: No
         - Analysis: The navigation target is inconsistent with the target object's location. The robot should first navigate to the office to pick up the book, or the object location should be changed to the kitchen.

       Example 4
       Input:
        {
          "robot_type": "dual-arm robot",
          "task_name": "stack_marker_on_pencil",
          "task_description": "The robot picks up a marker and stacks it on top of a pencil.",
          "objects": ["marker", "pencil"],
          "skills": ["pick", "stack"],
          "scene_layout": {
            "marker": "left side of table",
            "pencil": "center of table"
          },
          "task_context": "Both objects are cylindrical and the marker is expected to remain balanced on top of the pencil."
        }
       Output:
         - Feasibility: No
         - Analysis: The task is physically implausible because stacking one cylindrical object on another provides an unstable contact. Replace the target support with a flat object or change the task from stacking to placing the marker beside the pencil."
}
\end{lstlisting}
\end{tcolorbox}

\textbf{Self-Reflection Refiner Prompt}
The refiner synthesizes critiques to produce the improved task $T_{ref}$. The prompt includes:
(1) \textit{Context Aggregation:} Here is the original task $T_{raw}$ and the critiques from the evaluators.
(2) \textit{Memory Retrieval:} Consider these heuristic guidelines retrieved from historical execution failures: [Retrieved Knowledge $k$].
(3) \textit{Revision Instruction:} Modify $T_{raw}$ to resolve all identified issues while maintaining the original semantic intent.

\begin{tcolorbox}[
  title=Self-Reflection Refiner Prompt,
  colback=gray!10,
  colframe=gray!50,
  fonttitle=\bfseries,
  breakable,
  before skip=6pt,
  after skip=6pt
]

\begin{lstlisting}[
  basicstyle=\ttfamily\scriptsize,
  breaklines=true,
  columns=fullflexible,
  breakatwhitespace=false,
  keepspaces=true,
  escapeinside={(*}{*)}
]
{
  "(*\textbf{system}*)":
    "You are an expert in dual-arm robotic task design and refinement, with strong expertise in synthesizing feedback from multiple evaluators. Your responsibility is to generate a final, improved version of a dual-arm robotic task specification by integrating: 
       (1) the originally generated task,
       (2) feasibility and quality critiques provided by three independent reviewer LLMs.
     Your refinement must ensure that the resulting task is physically executable, logically coherent, and suitable for high-quality robotic policy learning.",

  "(*\textbf{user}*)":
    "You will be provided with structured task descriptions in a JSON-like format and the judgements and criteria from three critic models. Your task is to act as a task refiner to synthesize feedback from a 'Critic' module and the 'Original Task' definition to generate high-fidelity, physically plausible, and diverse dual-arm robotic manipulation tasks.

    1. Evaluation Criteria:
    
       You will be provided with a structured task description in a JSON-like format, along with evaluation results from three critic models.
       
       The critic models are:
       (1) Constraint Adherence Evaluator as a hard constraint
       (2) Physical Feasibility Evaluator as a hard constraint
       (3) Task Novelty Evaluator as a soft constraint

       You must process and resolve critic feedback in the following strict order:

       Step 1. Constraint Adherence Resolution:
         This step has the highest priority.
         If the Constraint Adherence Evaluator returns Feasibility: No, you must revise the task to ensure:
         - All object names exactly match entries in asset.csv
         - All skill names exactly match entries in Skills.txt
         - No hallucinated entities or properties are present
         Exact string matching is mandatory. You are not allowed to introduce new objects or skills beyond the provided candidate sets.

       Step 2. Physical Feasibility Resolution:
         This step is processed only after all constraint violations are resolved.
         If the Physical Feasibility Evaluator returns Feasibility: No, you must revise the task to satisfy:
         - Kinematic reachability
         - Stable contact and support conditions
         - Realistic dual-arm coordination requirements
         Prefer the following revision strategies in order:
         (a) Adjust spatial layout or object positions
         (b) Decompose or reorder sub-actions
         (c) Substitute objects only if absolutely necessary

       Step 3. Task Novelty Optimization:
         This step is applied only if the task passes Steps 1 and 2.
         If the Task Novelty Evaluator returns No, you may revise the task to improve diversity by:
         - Replacing overused skills with less frequent primitives from Skills.txt
         - Introducing multi-stage or tool-mediated interactions

       Step 4. Global Consistency Audit:
         After all revisions, ensure that:
         - The task is logically coherent
         - Object states, scene layout, and language instructions are mutually consistent
         - Dual-arm coordination is explicit and non-trivial

    2. Input Data Structure:
    
       (1) Initial tasks:
           Each task must be a JSON object containing:
           - 'task_name': Following the pattern 'RoboGene_robot_action'.
           - 'task_description': A detailed description.
           - 'language_instruction': A precise command for the VLA model.
           - 'objects': List of objects.
           - 'skills': List of skills.
           - 'scene_layout': 3x3 grids for left/right arm workspaces.
           - 'task_context': Explains physical state and positioning.

       (2) Feedback from three critic models:
           The input will be provided in the following format:
           - Feasibility: [Yes / No]
           - Analysis: [A concise justification grounded in evaluation dimensions]

    3. Output Format:
    
       The refined tasks will be provided in a JSON-like format and must include:
       - 'task_name', 'task_description', 'language_instruction', 'objects', 'skills', 'scene_layout', 'task_context'."
}
\end{lstlisting}

\end{tcolorbox}

\section{Generated Task Instances}
\label{app:task_instances}

To demonstrate the versatility of RoboGene, we present representative JSON outputs for single-arm, dual-arm, and mobile manipulation tasks. 
These examples illustrate how the system generates specific object layouts and skill sequences.

\paragraph{Instance 1: Single-Arm Task}
This long-horizon task involves a sequence of cleaning operations, requiring the robot to manage state changes (dirty to clean) and precise pick-and-place actions.

\begin{tcolorbox}[title=Single-Arm Task, colback=gray!10, colframe=gray!50, fonttitle=\bfseries, breakable, before skip=6pt, after skip=6pt]
\begin{lstlisting}[
    basicstyle=\ttfamily\scriptsize, % 缩小字体以适应单栏
    breaklines=true,                 % 开启自动换行
    columns=fullflexible,            % 紧凑排列，避免单词间空隙过大
    breakatwhitespace=false,         % 允许在任意字符处换行(防止长字符串溢出)
    keepspaces=true                  % 保持JSON缩进
]
{
  "task_name": "robogene_single_clean_spill",
  "task_description": "The robot uses a tissue to clean spilled drink under the display unit: first, throw the used tissue into the trash bin, then take a new tissue to wipe the table.",
  "language_instruction": "Clean the spilled drink under the display unit: throw the used tissue into the trash bin first, then wipe the table with a new tissue.",
  "objects": [
    "C&S Tissue Pack",
    "Countertop Display Unit",
    "Beverage Bottle",
    "Trash Bin"
  ],
  "skills": ["clean"],
  "scene_layout": {
    "C&S Tissue Pack": "Lower Right",
    "Countertop Display Unit": "Upper Center",
    "Beverage Bottle": "Toppled on the table beside the display unit",
    "Trash Bin": "Lower Right, near the tissue"
  },
  "task_context": "The used tissue is first deposited into the bin. Subsequently, a new tissue is retrieved to wipe the liquid on the desktop and the area beneath the shelf."
}
\end{lstlisting}
\end{tcolorbox}



\paragraph{Instance 2: Dual-Arm Task}
This task highlights the system's ability to generate bimanual coordination tasks where one arm acts as a stabilizer while the other performs a precision tool-use operation.

\begin{tcolorbox}[title= Dual-Arm Task, colback=gray!10, colframe=gray!50, fonttitle=\bfseries, breakable, before skip=6pt, after skip=6pt]
\begin{lstlisting}[
    basicstyle=\ttfamily\scriptsize, % 缩小字体以适应单栏
    breaklines=true,                 % 开启自动换行
    columns=fullflexible,            % 紧凑排列，避免单词间空隙过大
    breakatwhitespace=false,         % 允许在任意字符处换行(防止长字符串溢出)
    keepspaces=true                  % 保持JSON缩进
]
{
  "task_name": "robogene_dual_rotate_screwdriver",
  "task_description": "The left arm stabilizes a wooden block containing a large Phillips screw. The right arm grasps a screwdriver, aligns its tip with the screw head groove, inserts it, and rotates the wrist 90 degrees clockwise while applying downward pressure to tighten.",
  "language_instruction": "Hold the wooden block firmly with your left arm. Pick up the screwdriver with your right arm, insert its tip into the screw head, and then rotate it clockwise 90 degrees to tighten it.",
  "objects": [
    "Screwdriver", 
    "Wooden Block with Screw"
  ],
  "skills": ["rotate"],
  "scene_layout": {
    "Screwdriver": "Right Arm Workspace: Middle Right",
    "Wooden Block with Screw": "Left Arm Workspace: Middle Left"
  },
  "task_context": "The block is placed flat with the screw head facing up and loose. The screwdriver handle is oriented towards the right arm, placed horizontally."
}
\end{lstlisting}
\end{tcolorbox}

\paragraph{Instance 3: Mobile Manipulation Task}
This task necessitates navigation and logical reasoning to sort objects based on size, demonstrating spatial planning across a larger workspace.

\begin{tcolorbox}[title=Mobile Manipulation Task, colback=gray!10, colframe=gray!50, fonttitle=\bfseries, breakable, before skip=6pt, after skip=6pt]
\begin{lstlisting}[
    basicstyle=\ttfamily\scriptsize, % 缩小字体以适应单栏
    breaklines=true,                 % 开启自动换行
    columns=fullflexible,            % 紧凑排列，避免单词间空隙过大
    breakatwhitespace=false,         % 允许在任意字符处换行(防止长字符串溢出)
    keepspaces=true                  % 保持JSON缩进
]
{
  "task_name": "robogene_sort_items",
  "task_description": "Classify cardboard boxes of varying sizes and deposit them into the corresponding blue storage bins.",
  "language_instruction": "Sort the cardboard boxes by size and place them into the corresponding blue storage bins.",
  "objects": [
    "Cardboard Box (No White Edge) 4#",
    "Blue Storage Bins"
  ]   
  "skills": [
    "pick", 
    "place",
    "identify",
    "navigate"
  ],
  "scene_image": "industry_scenario.jpg",
  "scene_layout": {
    "Small Cardboard Box": "On conveyor belt, near left side",
    "Large Cardboard Box": "On conveyor belt, near right side",
    "Small Blue Storage Bins": "On the right side of the table",
    "Large Blue Storage Bins": "On the right side of the table",
  },
  "steps": [
    {
      "step": 1, 
      "skill": "pick", 
      "action": "Pick up a cardboard box", 
      "requirement": "Select a target box"
    },
    {
      "step": 2, 
      "skill": "identify", 
      "action": "Identify box size", 
      "requirement": "Classify based on dimensions."
    },
    {
      "step": 3, 
      "skill": "navigate", 
      "action": "Navigate to corresponding bin", 
      "requirement": "Move to correct storage location"
    },
    {
      "step": 4, 
      "skill": "place", 
      "action": "Place box in the blue storage bin", 
      "requirement": "Ensure stable placement"
    }
  ],
}
\end{lstlisting}
\end{tcolorbox}

\clearpage

\end{document}